\documentclass{article}

\PassOptionsToPackage{numbers}{natbib} 
 \usepackage[final,sglblindworkshop]{neurips_2025}
\usepackage[utf8]{inputenc} 
\usepackage[T1]{fontenc}    
\usepackage{hyperref}       
\usepackage{url}            
\usepackage{booktabs}       
\usepackage{amsfonts}       
\usepackage{nicefrac}       
\usepackage{microtype}      
\usepackage{xcolor}         
\usepackage{graphicx}
\usepackage{amsmath} 
\usepackage[linesnumbered,ruled,vlined]{algorithm2e}
\usepackage{amssymb}
\usepackage{subcaption}
\usepackage{booktabs}
\usepackage{bm}
\usepackage{mathtools}
\usepackage{amsthm}
\usepackage{cleveref}
\usepackage{breqn}
\usepackage{caption}
\usepackage{wrapfig}

\newcommand{\alg}[1]{Algorithm~#1}
\newcommand{\fig}[1]{Figure~#1}
\newcommand{\eq}[1]{Equation~#1}
\newcommand{\scn}[1]{Sec~#1}
\newcommand{\tbl}[1]{Table~#1}

\newcommand{\apx}[1]{Appendix~#1}

\title{How Should We Evaluate Data Deletion in Graph-Based ANN Indexes?}
\workshoptitle{Workshop on ML for Systems at NeurIPS 2025}

\author{%
  Tomohiro Yamashita \\
  The University of Tokyo\\
  \texttt{t\_yamashita@hal.t.u-tokyo.ac.jp}
  \And
  Daichi Amagata \\
  The University of Osaka\\
  \texttt{amagata.daichi@ist.osaka-u.ac.jp}
  \And
  Yusuke Matsui \\
  The University of Tokyo \\
  \texttt{matsui@hal.t.u-tokyo.ac.jp}
}

\begin{document}

\maketitle

\begin{abstract}
Approximate Nearest Neighbor Search (ANNS) has recently gained significant attention due to its many applications, such as Retrieval-Augmented Generation.
Such applications require ANNS algorithms that support dynamic data, so the ANNS problem on dynamic data has attracted considerable interest.
However, a comprehensive evaluation methodology for data deletion in ANNS has yet to be established.
This study proposes an experimental framework and comprehensive evaluation metrics to assess the efficiency of data deletion for ANNS indexes under practical use cases.
Specifically, we categorize data deletion methods in graph-based ANNS into three approaches and formalize them mathematically. 
The performance is assessed in terms of accuracy, query speed, and other relevant metrics.
Finally, we apply the proposed evaluation framework to Hierarchical Navigable Small World, one of the state-of-the-art ANNS methods, to analyze the effects of data deletion, and propose Deletion Control, a method which dynamically selects the appropriate deletion method under a required search accuracy.
\end{abstract}

\section{Introduction}
\label{sec:intro}
ANNS is an essential building block for applications such as Retrieval-Augmented Generation (RAG)~\cite{rag} and recommendation systems~\cite{recommend}.
In these applications, frequent data updates occur due to the addition of new products and the removal of unavailable items.  
Consequently, research has been conducted on ANNS algorithms that support data deletion, including IVF-based methods \cite{ada-ivf}, product quantization-based methods \cite{turbo-lvq}, and graph-based methods \cite{freshdiskann,hnsw-update,ipgm}.

However, no comprehensive methodology has been established for evaluating data deletions under practical use cases.  
Evaluating the execution time of data deletion and the search performance after deletion is crucial for selecting ANNS algorithms.
Moreover, the experimental settings in existing evaluation are unrealistic, such as re-adding deleted data~\cite{freshdiskann,hnsw-update}.
The evaluation criteria used in existing studies, furthermore, are not sufficiently comprehensive for assessing deletion performance~\cite{ada-ivf,ipgm,spfresh}.  

To address these limitations, we develop a unified, deployment-oriented evaluation methodology for data deletion in graph-based ANNS. We first formalize three deletion methods (logical deletion, physical deletion, and rebuilding) and implement them within Hierarchical Navigable Small World (HNSW)~\cite{hnsw}. We then introduce an experimental protocol that measures deletion latency, search/insert throughput, memory footprint, and post-deletion accuracy under realistic workloads.
Based on the experimental results, we propose an algorithm that dynamically switches between deletion methods.

\section{Related Work}
\label{sec:related-work}

Ada-IVF~\cite{ada-ivf} and SPFresh~\cite{spfresh}, which employ the inverted file structure, handle data updates by reassigning vectors to clusters.  
The following methods in graph-based ANNS algorithms have implemented support for data deletion.  
First, FreshDiskANN~\cite{freshdiskann} is an algorithm based on DiskANN~\cite{diskann}, which utilizes disk storage.
It maintains search accuracy by reconnecting edges among neighboring nodes when a node is deleted.  
Additionally, the MN-RU algorithm \cite{hnsw-update} addresses the issue of unreachable nodes in HNSW by maintaining a backup graph to preserve search accuracy after data deletions.  
Furthermore, the IPGM algorithm \cite{ipgm} recalculates all neighboring nodes within one hop of a deleted node and reconnects edges to maintain search accuracy.  
Existing studies have lacked a practical experimental setup. In this work, we propose a realistic and effective evaluation methodology.

\section{Formal Definition of Deletion}
\label{sec:approach}
We define baseline data deletion methods in graph-based ANNS using pseudocode.  
Specifically, we categorize these methods into three types: logical deletion, physical deletion, and rebuilding.  

\begin{wrapfigure}[6]{r}{0.48\textwidth} 
  \vspace{-\baselineskip} 
  \begin{algorithm}[H]
    \footnotesize
    \caption{Search in logical deletion}
    \label{alg:logical-delete}
    \KwIn{$\mathbf{q}$, $\mathcal{P}$, $\mathcal{N}$, $\mathcal{F}$\\}
    \KwOut{$\mathcal{R}$}
    $\mathcal{R} \leftarrow \texttt{SEARCH}(\mathbf{q},\mathcal{P},\mathcal{N}) \setminus \mathcal{F}$ \\
    \textbf{return} $\mathcal{R}$
  \end{algorithm}
\end{wrapfigure}

\paragraph{Preliminaries}
\label{sec:preparation}

Let $\mathcal{P} = \{\mathbf{p}_i\}^n_{i=1} \subset \mathbb{R}^d$ be the set of $n$ nodes in the graph, where each node $\mathbf{p}_i \in \mathbb{R}^d$ is a $d$-dimensional vector. 
For each node $\mathbf{p}_i$, let $\mathcal{N}_i \subset \{1,2,\dots,n\}$ be the set of indices of its neighboring nodes.  
We define $\mathcal{N} = \{\mathcal{N}_i\}^n_{i=1}$ as the collection of all neighborhood sets.  
Thus, the graph is represented by $\mathcal{P}$ and $\mathcal{N}$.

\begin{wrapfigure}[12]{r}{0.48\textwidth}
  \vspace{-\baselineskip}
  \begin{algorithm}[H]
    \footnotesize
    \caption{Physical deletion}
    \label{alg:physical-delete}
    \KwIn{$\mathcal{D}$, $\mathcal{P}$, $\mathcal{N}$\\}
    \KwOut{$\mathcal{P'}$, $\mathcal{N'}$\\}
    \BlankLine
    \ForEach{$\mathbf{p}_i \in \mathcal{P}$} {
      \eIf{$i \in \mathcal{D}$}{
        $\mathcal{P} \leftarrow \mathcal{P} \setminus \{\mathbf{p}_i\}$\\
        $\mathcal{N} \leftarrow \{\mathcal{N}_j \in \mathcal{N} \mid j \ne i\}$
      }{
        $\mathcal{N}_i \leftarrow \mathcal{N}_i \setminus \mathcal{D}$
      }
    }
    \textbf{return} $\mathcal{P}, \mathcal{N}$
  \end{algorithm}
\end{wrapfigure}

We define the standard algorithms for search and graph construction, shared across all deletion approaches.  
Let $\mathbf{q} \in \mathbb{R}^d$ be a query vector.
We define the search algorithm as \texttt{SEARCH}$: (\mathbf{q},\mathcal{P},\mathcal{N}) \mapsto \mathcal{R}$, which takes the query vector and the graph as input and returns a set of approximate nearest neighbor $\mathcal{R} \subset \{1,2,\dots,n\}$.  
Similarly, we define the graph construction algorithm as \texttt{CONSTRUCT}$: \mathcal{P} \mapsto \mathcal{N}$, which takes the set of nodes $\mathcal{P}$ as input and returns the neighbor set $\mathcal{N}$.  
Finally, let $\mathcal{D} \subset \{1,2,\dots,n\}$ denote the set of node indices corresponding to deletion queries.  
We define the deletion algorithm as \texttt{DELETE}$: (\mathcal{D},\mathcal{P},\mathcal{N}) \mapsto (\mathcal{P'},\mathcal{N'})$, which takes $\mathcal{D}$, $\mathcal{P}$, and $\mathcal{N}$ as input and outputs the updated node set $\mathcal{P'} \subset \mathcal{P}$ and the updated neighborhood set $\mathcal{N'} \subset \mathcal{N}$ after deletion.  

\begin{wrapfigure}[6]{r}{0.48\textwidth}
  \vspace{-\baselineskip}
  \begin{algorithm}[H]
    \footnotesize
    \caption{Rebuilding}
    \label{alg:reconstruct-delete}
    \KwIn{$\mathcal{D}$, $\mathcal{P}$\\}
    \KwOut{$\mathcal{P'}$, $\mathcal{N'}$\\}
    \BlankLine
    $\mathcal{P} \gets \{\mathbf{p}_i \in \mathcal{P} \mid i \notin \mathcal{D} \}$ \\
    $\mathcal{N} \leftarrow \texttt{CONSTRUCT}(\mathcal{P})$ \\
    \textbf{return} $\mathcal{P}, \mathcal{N}$
  \end{algorithm}
\end{wrapfigure}

\begin{figure*}[tb]
    \centering
    \begin{subfigure}{0.30\textwidth}
        \centering
        \includegraphics[width=\textwidth]{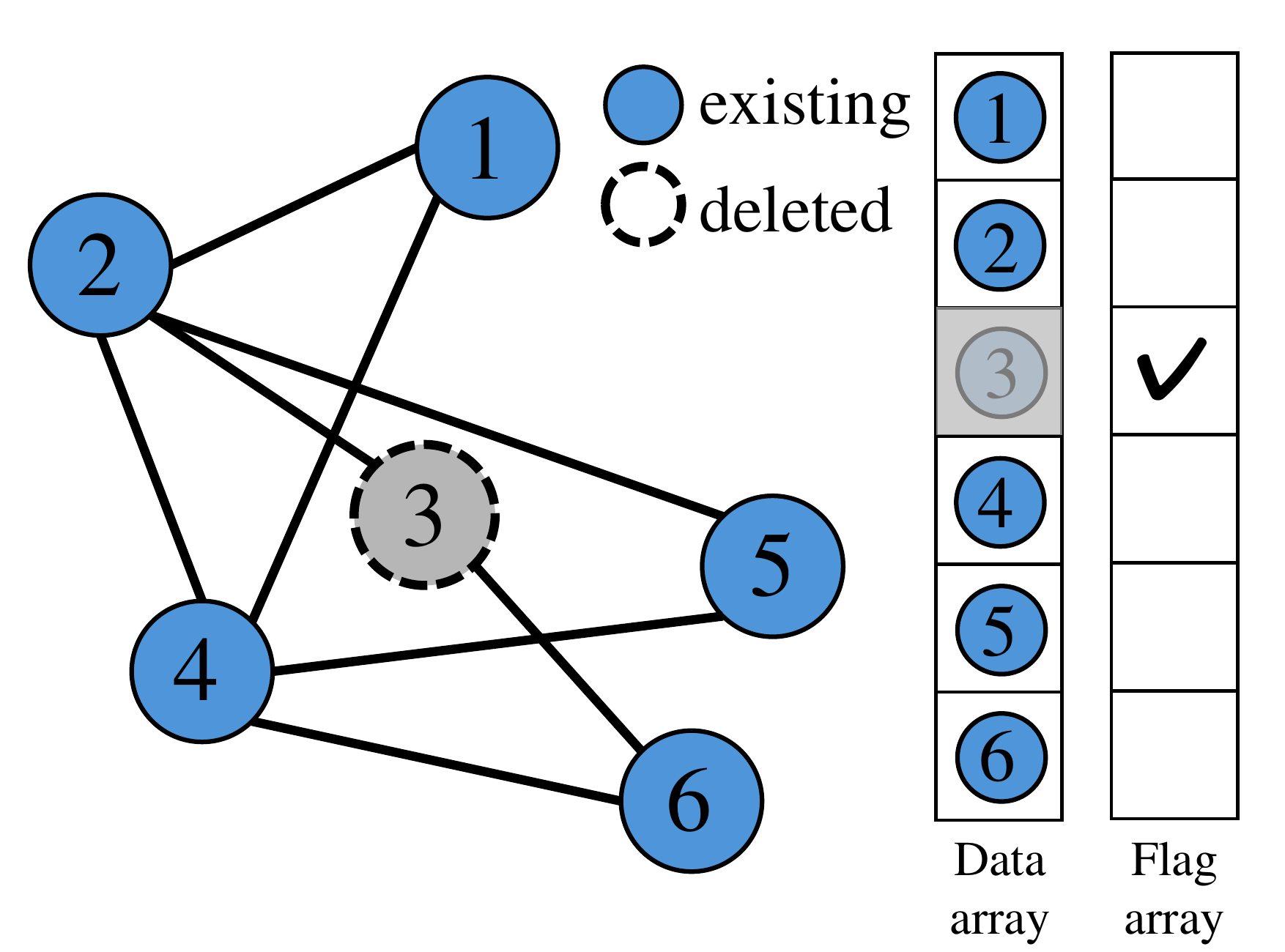}
        \caption{Logical deletion}
        \label{fig:logical-delete}
    \end{subfigure}
    \hfill
    \begin{subfigure}{0.30\textwidth}
        \centering
        \includegraphics[width=\textwidth]{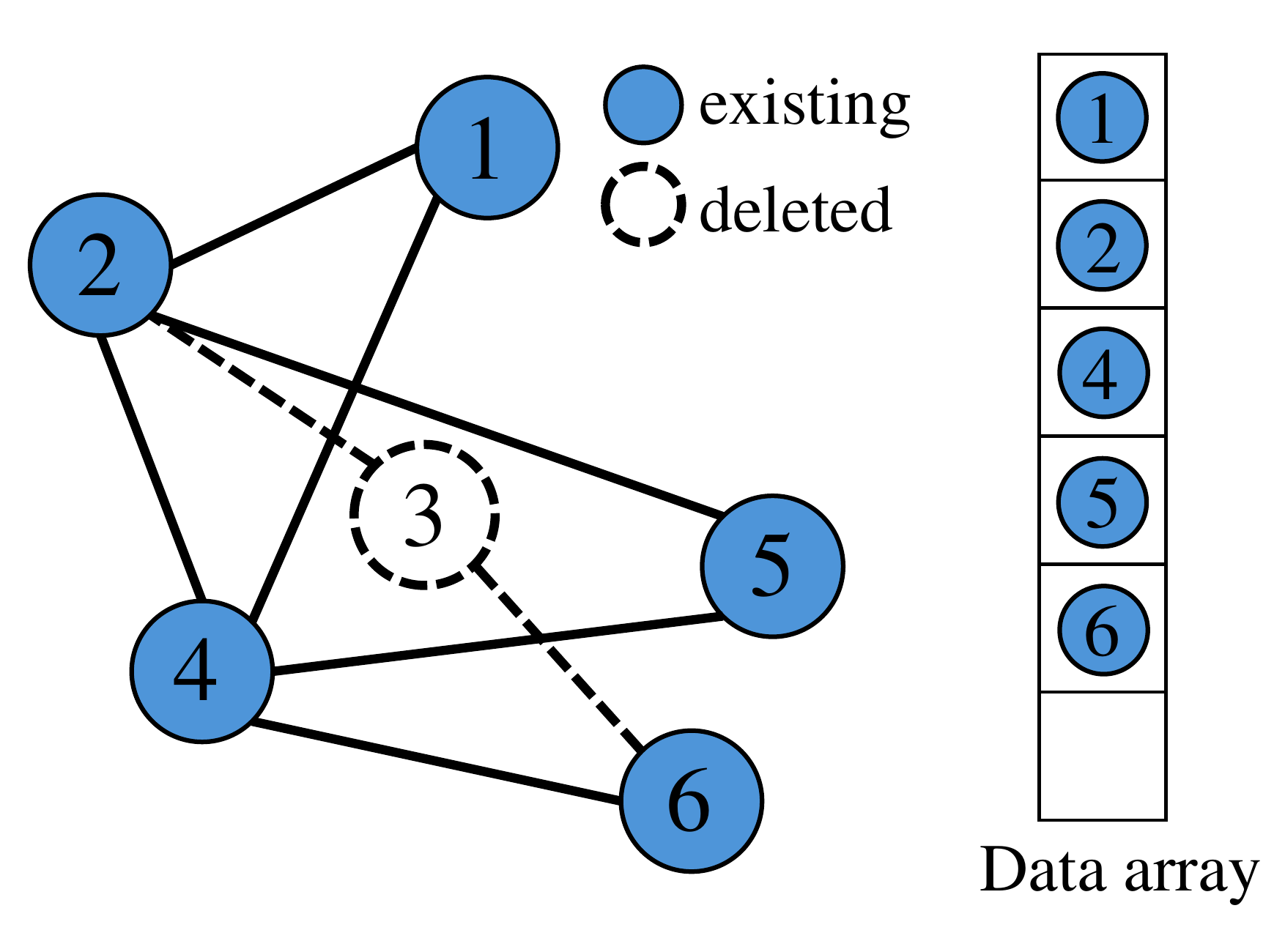}
        \caption{Physical deletion}
        \label{fig:physical-delete}
    \end{subfigure}
    \hfill
    \begin{subfigure}{0.30\textwidth}
        \centering
        \includegraphics[width=\textwidth]{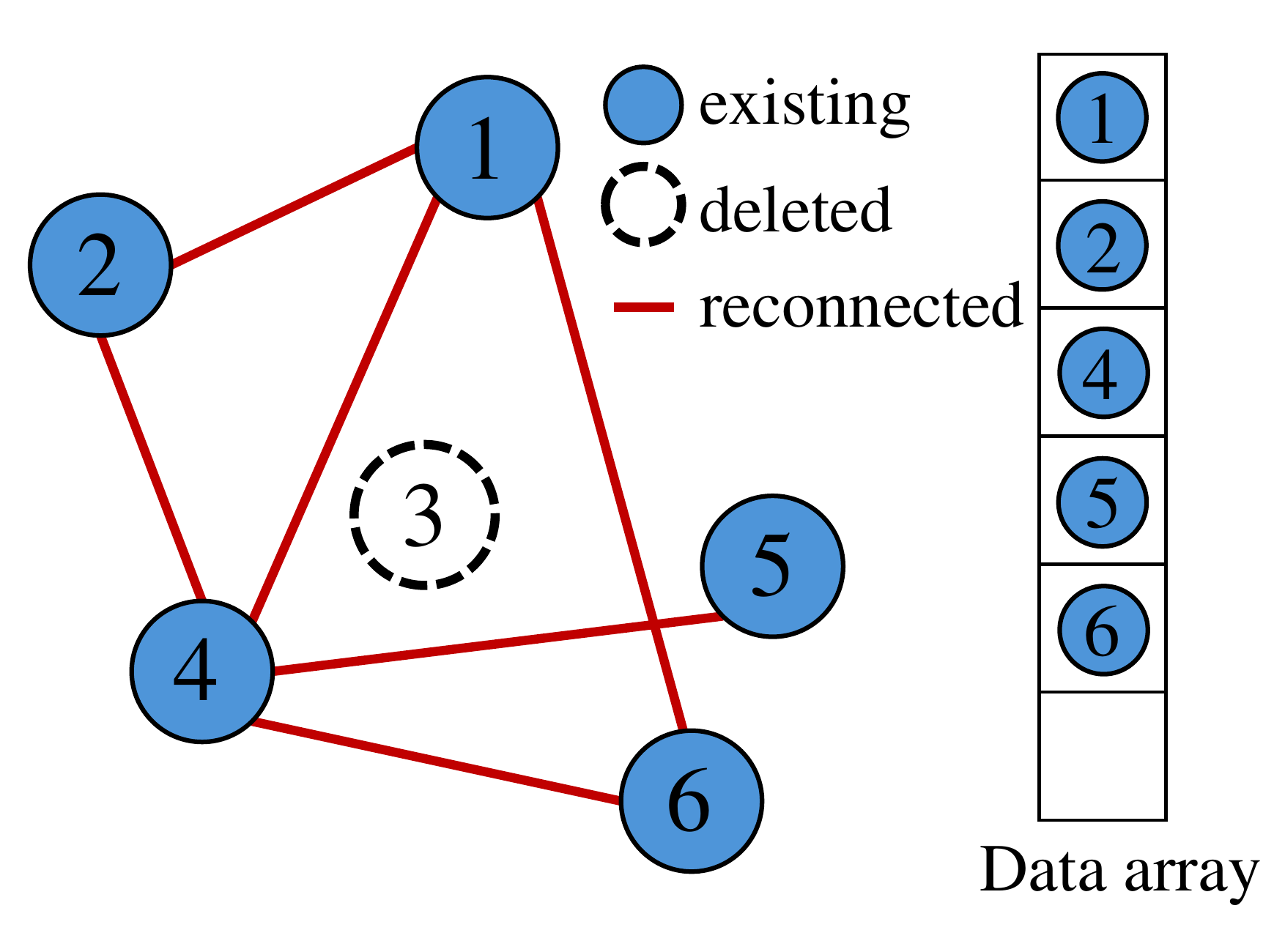}
        \caption{Rebuilding}
        \label{fig:reconstruct-delete}
    \end{subfigure}
    \hfill
    \caption{
    Overview of three data deletion methods: logical deletion, physical deletion, and rebuilding.
    }
    \label{fig:delete}
\end{figure*}

\paragraph{Logical Deletion}
\label{subsec:logical-delete}
Logical deletion is a method where each deleted node is marked with a flag $\mathcal{F} \subset \{1,2,\dots,n\}$ at the time of deletion.
Such flags are referenced during the search to exclude flagged nodes from the results.  
\fig{\ref{fig:logical-delete}} illustrates the mechanism of logical deletion in a graph.
The search algorithm is presented in \alg{\ref{alg:logical-delete}}.

\paragraph{Physical deletion}
\label{subsec:physical-delete}
Physical deletion is a method that removes the designated data by deleting all edges connected to the node, as illustrated in \fig{\ref{fig:physical-delete}}.  
The procedure for this deletion approach is presented in \alg{\ref{alg:physical-delete}}.  
The data is also removed from memory.

\paragraph{Rebuilding}
\label{subsec:rebuilding-delete}

Rebuilding is a method that removes data both from the graph by reconstructing the graph using all remaining data, as illustrated in \fig{\ref{fig:reconstruct-delete}}.  
The procedure for this deletion approach is presented in \alg{\ref{alg:reconstruct-delete}}.  
Similar to the physical deletion approach, this approach involves removing data from memory during the deletion process.  

More detailed explanation of each data deletion method are provided in \apx{\scn{\ref{sec:deletion-method-detail}}}.

\section{Experiments}
\label{sec:experiment}

\paragraph{Settings}
\label{subsec:settings}

For the experiments, we used SIFT1M~\cite{sift1m}, GIST1M~\cite{sift1m}, SIFT1B~\cite{sift1b}, DEEP1M~\cite{deep1m}, and Glove100Angular~\cite{glove}.  
For SIFT1B, we created a subset of $2\times10^6$ data points.  
We evaluated the performance by repeatedly performing insertion and deletion with the same batch size in the database.
The detailed experimental settings are described in \apx{\scn{\ref{sec:setting}}}.

\begin{figure*}[tb]
    \centering
    \begin{subfigure}{0.24\textwidth}
        \centering
        \includegraphics[width=\textwidth]{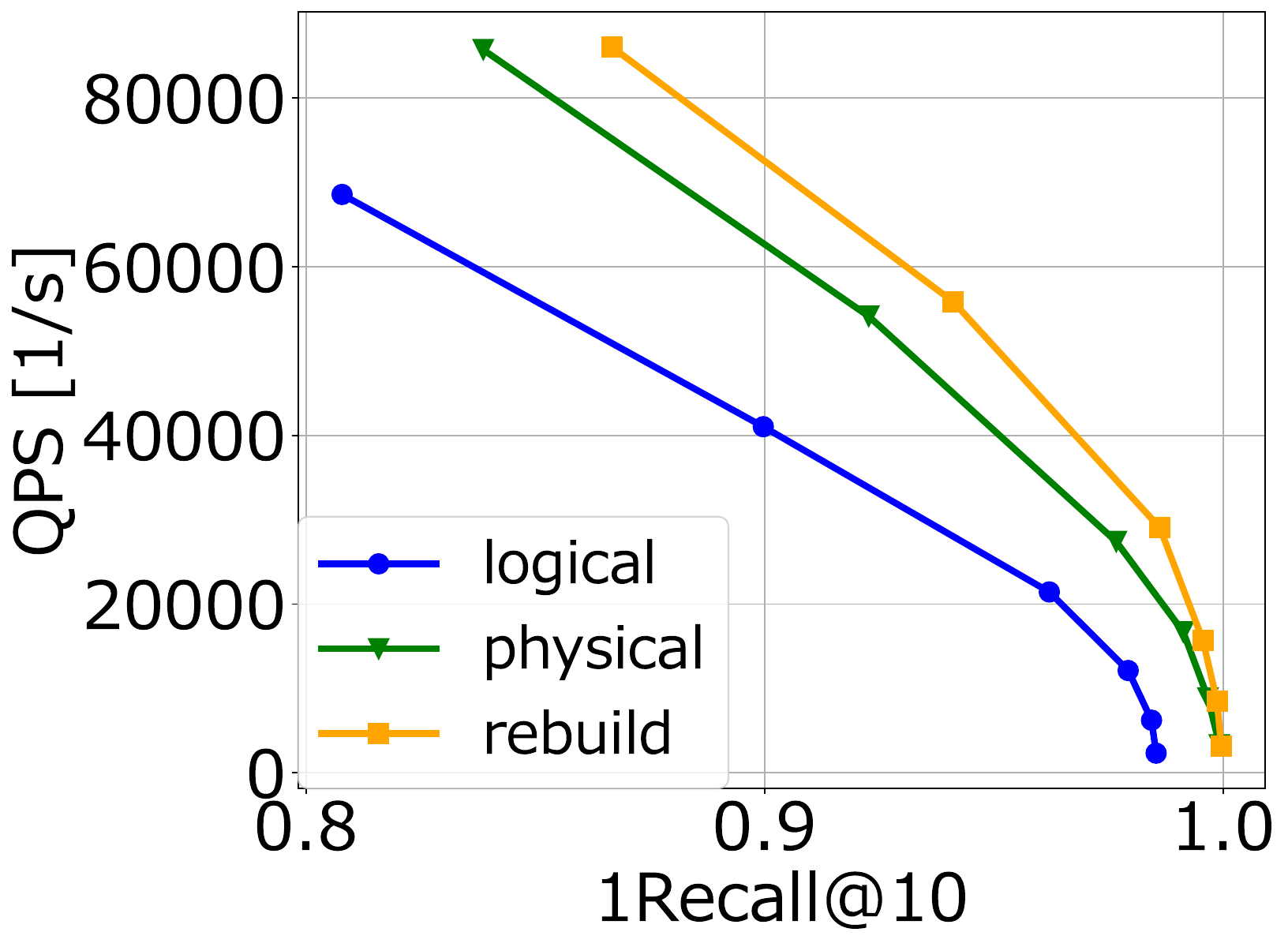}
        \caption{\scriptsize SIFT1M: Comparison of deletion methods at step 5}
        \label{fig:sift1m-qps-recall-step5-100000}
    \end{subfigure}
    \hfill
    \begin{subfigure}{0.24\textwidth}
        \centering
        \includegraphics[width=\textwidth]{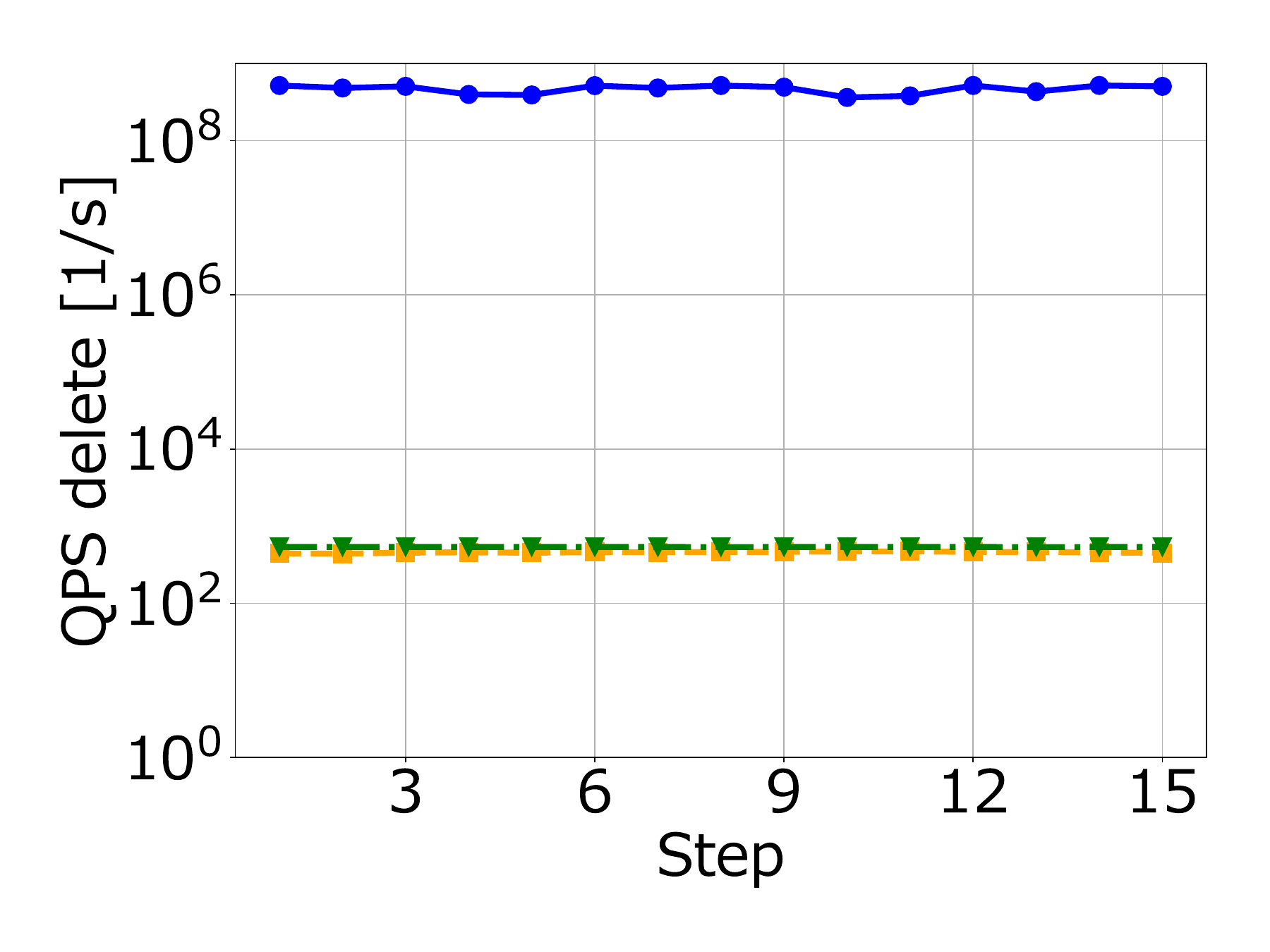}
        \caption{\scriptsize SIFT1B: QPS of deletion ($b = 10^5$)}
        \label{fig:sift1b2m-qps-delete-100000}
    \end{subfigure}
    \hfill
    \begin{subfigure}{0.24\textwidth}
        \centering
        \includegraphics[width=\textwidth]{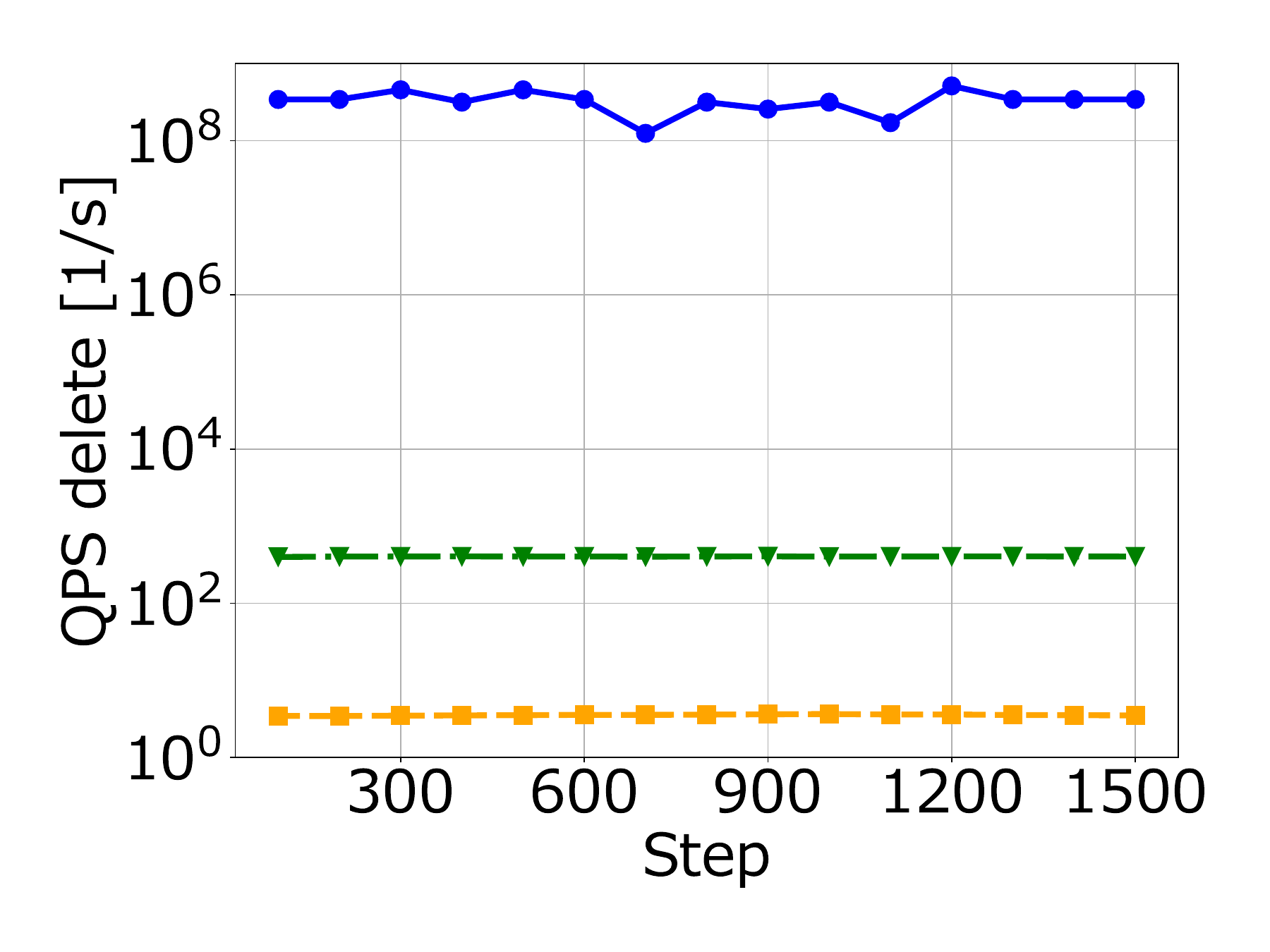}
        \caption{\scriptsize SIFT1B: QPS of deletion ($b = 10^3$)}
        \label{fig:sift1b2m-qps-delete-1000}
    \end{subfigure}
    \hfill
    \begin{subfigure}{0.24\textwidth}
        \centering
        \includegraphics[width=\textwidth]{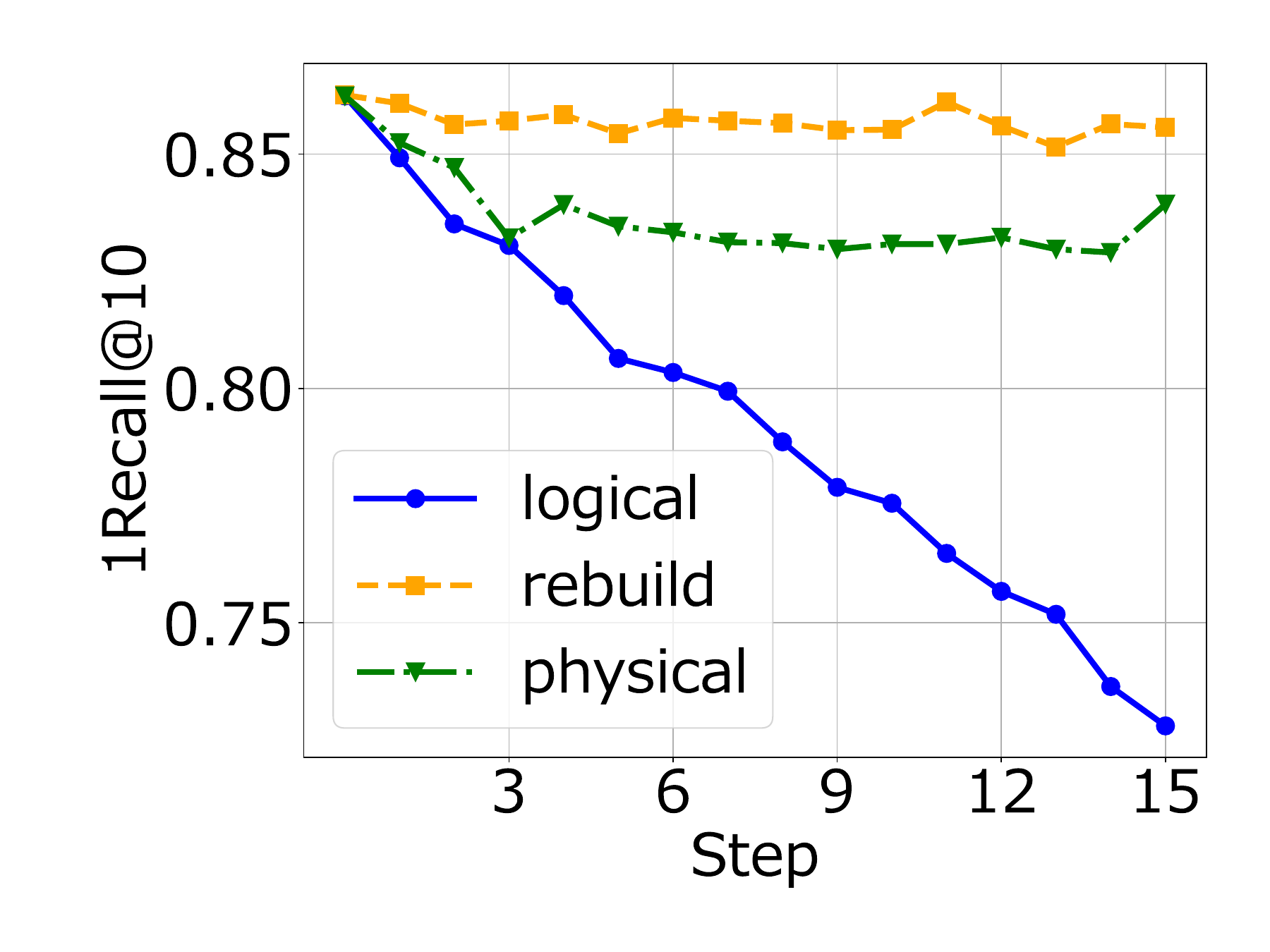}
        \caption{\scriptsize SIFT1B: 1-Recall@10 after deletion ($b = 10^5$)}
        \label{fig:sift1b2m-recall-100000}
    \end{subfigure}
    \hfill
    \caption{Evaluation of deletion performance on SIFT1M and SIFT1B datasets. (a) Comparison of search performance after step 5 deletion in SIFT1M. (b, c) QPS for deletion on SIFT1B with varying batch sizes. (d) Impact on 1-Recall@10 after updates on SIFT1B.}
    \label{fig:Evaluation}
\end{figure*}

\paragraph{Metrics}
\label{subsec:metrics}



We evaluated search accuracy using 1-Recall@10.  
To assess query processing speed, we used Queries Per Second (QPS).
We measured QPS-search for search speed evaluation, QPS-add for insertion speed evaluation, and QPS-delete for deletion speed evaluation.  
We adopt the QPS-Recall curve as an evaluation metric by plotting QPS-search against 1-Recall@10.

\paragraph{Experimental Results}
\label{subsec:experimental-results}

\fig{\ref{fig:Evaluation}} presents a comparison of search performance before and after data updates for each of the three deletion approaches.  
\fig{\ref{fig:sift1m-qps-recall-step5-100000}} shows that post-update search performance is highest with rebuilding, followed by physical deletion, and then logical deletion.
From \fig{\ref{fig:sift1b2m-qps-delete-100000}} and \fig{\ref{fig:sift1b2m-qps-delete-1000}}, it is evident that logical deletion achieves the highest data deletion speed.
Additionally, under frequent small-batch deletions, rebuilding is relatively slower than physical deletion.
\fig{\ref{fig:sift1b2m-recall-100000}} indicates that with repeated data updates, the search accuracy of logical deletion deteriorates.  
Interestingly, this figure suggests that in physical deletion, search accuracy stabilizes to a constant value after multiple updates.  
The results for the other metrics are presented in \apx{\scn{\ref{sec:all-experimental-results}}}.



\section{Deletion Control}
\label{sec:two-threshold-deletion}

\paragraph{Problem Statement}
\label{subsec:problem-setting}

Based on \scn{\ref{sec:experiment}}, we discuss how to control data deletion methods in scenarios that require high search accuracy under continuous deletions. 
Following the experimental setup in \scn{\ref{sec:experiment}}, we assume a situation where data updates are repeatedly performed with the same batch size $b$.
Here, the input consists of the dataset for retrieval, the accuracy target $\alpha \in (0,1]$, and a small query training set (query data with known ground truth). In this setting, we repeatedly delete data with batch size $b$. Our goal is to present a hybrid data deletion strategy such that, even after deletion, the search accuracy remains higher than $\alpha$.

\begin{wrapfigure}[12]{r}{0.35\textwidth} 
    \vspace{-\baselineskip} 
    \centering
    \includegraphics[width=0.95\linewidth]{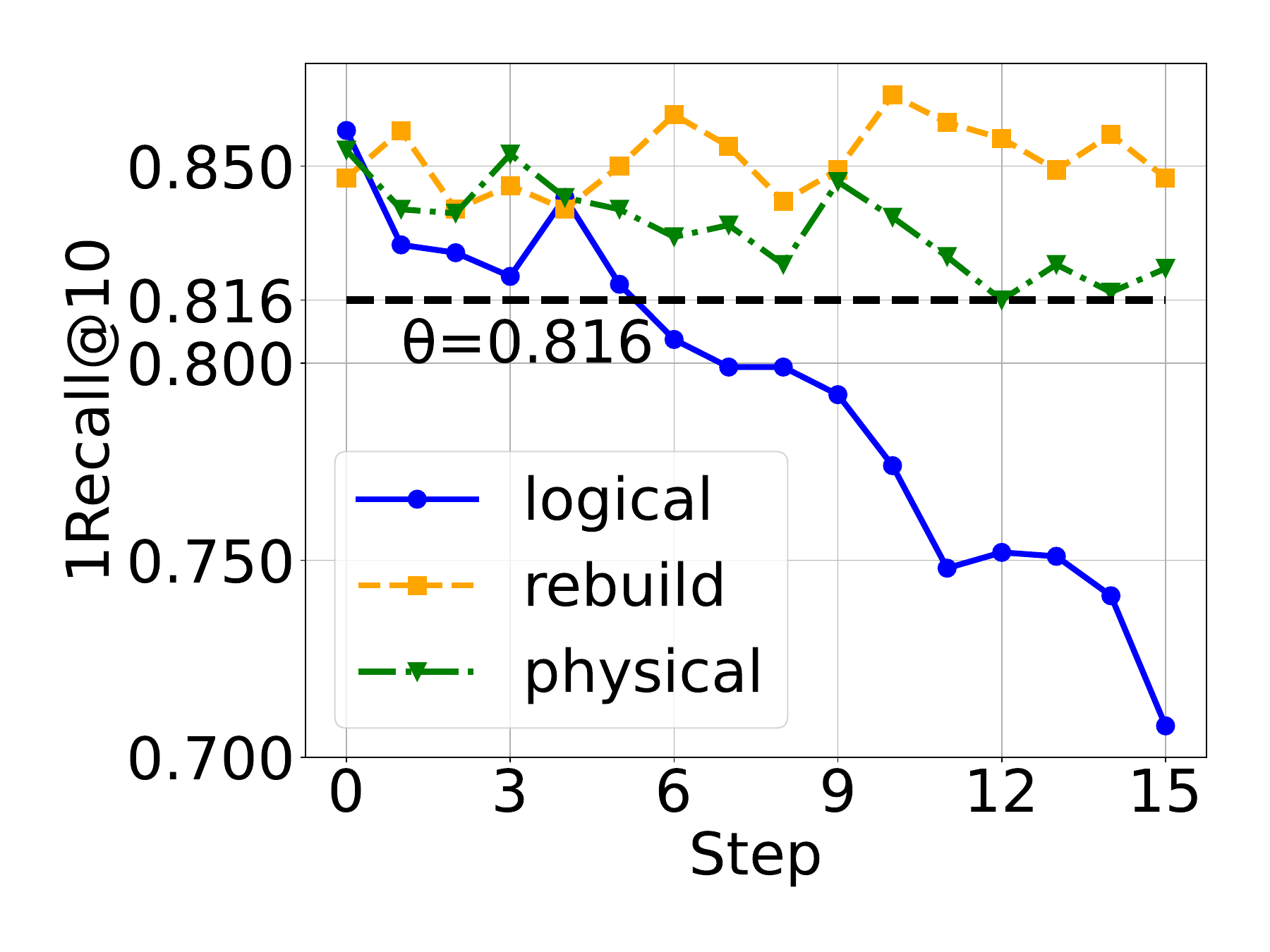}
    \caption{Comparison of 1-Recall@10 on SIFT1B training set (10$\%$ queries).}
    \label{fig:sift1b2m-recall-100000-10query}
\end{wrapfigure}

We consider two approaches. The first applies only physical deletion, while the second employs logical deletion and performs rebuilding once the performance begins to degrade.
The first approach is applicable when the requirement for search accuracy is not very strict.
In this case, continuing physical deletion does not reduce performance below the required level.
The second approach is employed when high search accuracy is required.
As shown in \fig{\ref{fig:sift1b2m-recall-100000}}, performance can be maintained as long as rebuilding is executed sufficiently often.
However, it is computationally expensive.
Therefore, logical deletion is applied until just before the performance drops below the required level, at which point rebuilding is performed.

\begin{wrapfigure}[13]{r}{0.35\textwidth}
    \vspace{-\baselineskip}
    \centering
    \includegraphics[width=0.95\linewidth]{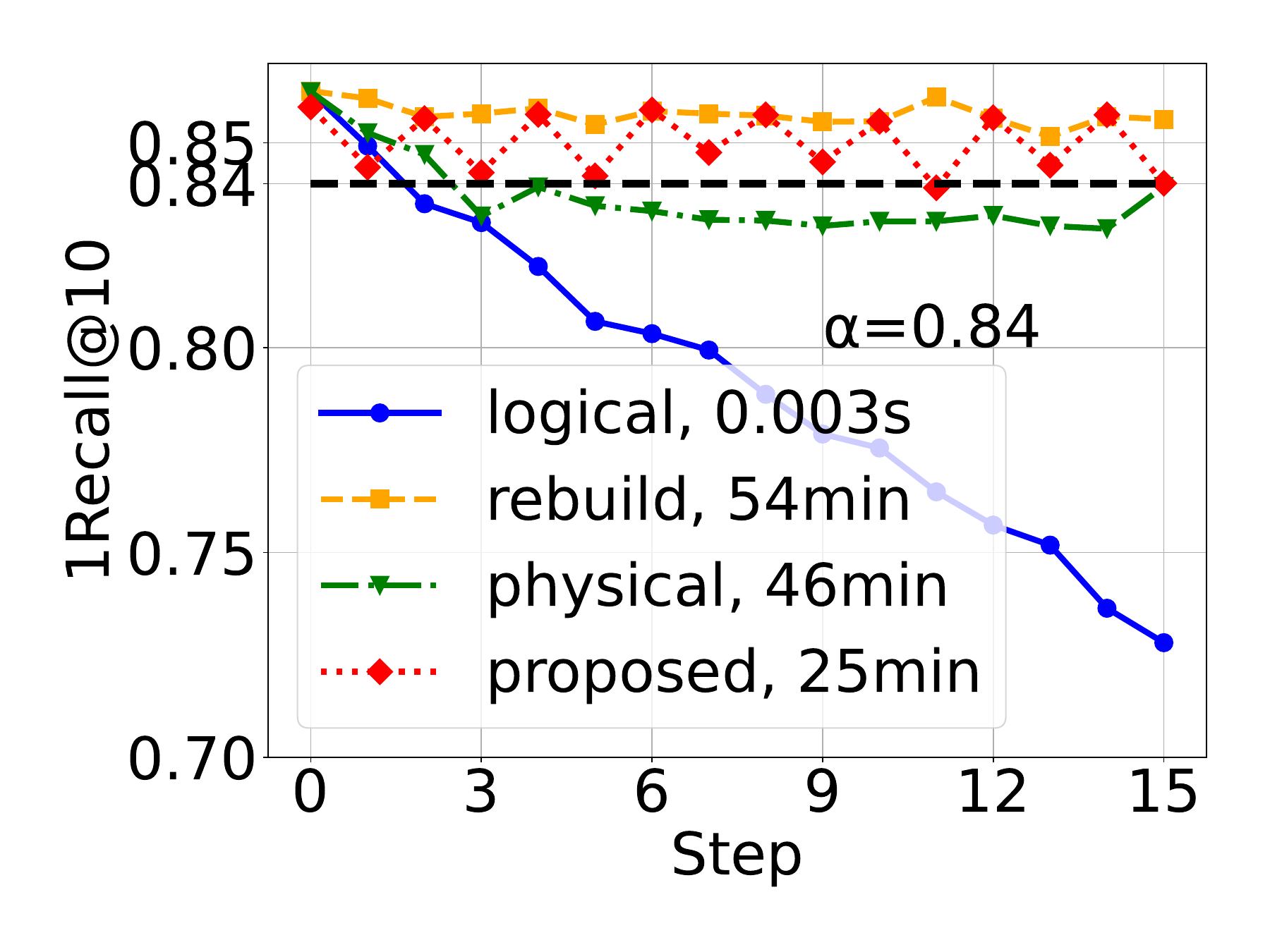}
    \caption{Comparison of 1-Recall@10 on test set (all queries).}
    \label{fig:sift1b2m-recall-100000-proposed}
\end{wrapfigure}

Here, we introduce two parameters, $\theta$ and $\pi$, necessary for designing the Deletion Control algorithm.
First, suppose we repeatedly perform only physical data deletion. Let us consider the minimum 1-Recall@10 achieved in this case and denote it by $\theta$. Next, we analyze the case where we repeatedly delete data logically.
We define $R_s \in (0,1]$ as the 1-Recall@10 after performing $s$ updates by the logical deletion, where $R_0$ is the 1-Recall@10 before deletion.
Here, we define $\pi$ as the maximum number of steps for which $R_s$ remains above $\alpha$:
\begin{equation}
    \label{eq:pi}
    \pi = \max_{R_s \ge \alpha} s.
\end{equation}
Here, $\theta$ and $\pi$ represent the dataset's characteristics, and cannot be measured unless actual query data is available.

\paragraph{Deletion Control Policy}
\label{subsec:selecting-deletion}

We first estimate $\theta$ and $\pi$ through experiments using the training set. 
As shown in \fig{\ref{fig:sift1b2m-recall-100000}}, the search accuracy of physical deletion converges to a stable value. 
Therefore, we can estimate $\theta$ as the lowest value of the measured 1-Recall@10.
Also, as shown in \fig{\ref{fig:sift1b2m-recall-100000}}, we can approximate $R_s$ as a linear function of the update step $s$. 
We therefore define $\Delta = (R_S - R_0)/S$ as the average decrease in 1-Recall@10 per step, where $S$ represents the maximum number of steps for the training set.
The $\pi$ can be estimated as $\pi \approx (\alpha - R_0)/\Delta$.

Then, we select one of the two deletion strategies.  
First, when $\alpha < \theta$, only physical deletion is repeatedly applied.
Since the 1-Recall@10 obtained by physical deletion never falls below $\theta$, the requirement is satisfied as $\alpha < \theta \le R_{s}$.
Second, when $\alpha \ge \theta$, the procedure alternates between performing logical deletion for $\pi$ steps and then executing one rebuilding operation. 
According to \eq{\ref{eq:pi}}, the condition $\alpha \le R_{s}$ holds for up to $\pi$ steps. 
Furthermore, as shown in \fig{\ref{fig:sift1b2m-recall-100000}}, rebuilding afterward restores the 1-Recall@10 to $R_{0}$. 
In this manner, the condition $\alpha \le R_{s}$ is consistently maintained, thereby satisfying the requirement.
The algorithm is shown in \apx{\alg{\ref{alg:deletion-control}}}.

\paragraph{Experiment}
\label{subsec:expriment-deletion-control}
We conducted an experiment on SIFT1B~\cite{sift1b} with $b = 10^5$, setting $\alpha = 0.84$. 
\fig{\ref{fig:sift1b2m-recall-100000-10query}} shows the 1-Recall@10 for the basic deletion methods using 10\% of the queries as the training set.
From these results, we estimate $\theta = 0.816$ and $\pi = 1$.
Since $\alpha > \theta$, we select the deletion strategy that alternates between logical deletion and rebuilding.
\fig{\ref{fig:sift1b2m-recall-100000-proposed}} indicates that the proposed method almost satisfies the required search accuracy. 
Furthermore, the proposed method has the smallest total deletion time among the deletion strategies that meet the accuracy requirement.


\section{Conclusion}
\label{sec:conclusion}
This study has three main contributions.  
First, we formally defined three baseline data deletion approaches for ANNS---logical deletion, physical deletion, and rebuilding---along with their mathematical formulations.  
Second, we established an experimental setup and evaluation metrics that align with practical use cases.  
Third, we implemented and empirically evaluated the baseline data deletion approaches on HNSW. We also proposed a deletion control algorithm that selects an appropriate data deletion method based on the required accuracy.

\textbf{Acknowledgements}: 
This work was supported by JST AIP Acceleration Research JPMJCR23U2, Japan.

\newpage

\bibliographystyle{unsrtnat}  
\bibliography{reference}

\newpage


\appendix\label{appendix}

\section{Algorithm of three deletion methods}
\label{sec:deletion-method-detail}
\subsection{Logical deletion}
Let $\mathcal{F} \subset \{1,2,\dots,n\}$ be a set of flags.
When performing deletion, the deletion flag set $\mathcal{F}$ is updated for the deletion query set $\mathcal{D}$ as follows:  
\begin{equation}
\label{eq:logical-delete}
    \mathcal{F} \leftarrow \mathcal{F} \cup \mathcal{D}
\end{equation}  

During the search process, nearest neighbor candidates are obtained by excluding deleted data from the initial results.  
The search algorithm incorporating this approach is presented in \alg{\ref{alg:logical-delete}}.  
Since logical deletion removes data only from the search results, the effect of data deletion appears at search time.  

Since logical deletion only involves updating flags, it can be performed efficiently.  
However, because the deleted nodes remain in $\mathcal{P}$ and the neighborhood set $\mathcal{N}$, memory consumption accumulates over time.  
Additionally, if insertions and deletions are repeated, the index has fewer not-deleted vectors.   
The operation that excludes deleted data from the initial search results may eventually lead to an empty result.

\subsection{Physical deletion}
During the search for a query vector $\mathbf{q}$, a straightforward search is performed following \eq{\ref{eq:search-nomal}}.  
\begin{equation}
\label{eq:search-nomal}
    \mathcal{R} \leftarrow \texttt{SEARCH}(\mathbf{q},\mathcal{P},\mathcal{N})
\end{equation}

Unlike logical deletion, physical deletion actually removes data and does not require index rebuilding. As a result, it does not need to retain vector data for distance calculations, making it the most memory-efficient method when implemented properly.    
However, the effectiveness of data removal heavily depends on the implementation of the ANNS algorithm and the memory layout.  
Furthermore, since deleting edges alters the structure of the graph, it is expected to affect search performance.  

\subsection{Rebuilding}
Since \texttt{CONSTRUCT} is invoked whenever a new update batch is given and distance computations are performed, the set of graph nodes $\mathcal{P}$ is required.  
On the other hand, reconstructing the index ensures that its structure remains optimal, preventing search performance degradation while maintaining appropriate memory consumption.





\section{Experimental Settings}
\label{sec:setting}
\begin{figure}[tb]
    \hfill
    \begin{minipage}[t]{0.97\textwidth}
        \centering
        \captionsetup{type=table}            
        \captionof{table}{Experimental conditions for each dataset}
        \label{table:experiment}
        \vskip 0.05in
        \begin{small}
        \begin{tabular}{@{}lcccc@{}}
        \toprule
        Dataset & $d$ & $n_o$ & $n$ & $b$\\
        \midrule
        SIFT 1M~\cite{sift1m} & $128$ & $ 10^6$& $5 \times 10^5$ & $10^5$ \\
        GIST 1M~\cite{sift1m} & $960$ & $ 10^6$& $5 \times 10^5$ & $10^5$ \\
        SIFT 1B~\cite{sift1b} & $128$ & $2 \times 10^6$& $5 \times 10^5$ & $10^3$ or $10^5$ \\
        DEEP 1M~\cite{deep1m} & $96$ & $ 10^6$& $5 \times 10^5$ & $10^5$ \\
        Glove100Angular~\cite{glove} & $100$ & $ 10^6$& $5 \times 10^5$ & $10^5$ \\
        \bottomrule
        \end{tabular}
        \end{small}
    \end{minipage}
\end{figure}

The datasets in~\tbl{\ref{table:experiment}} were partitioned for insertion and deletion operations, and the ground-truth is recomputed.  
Let the original dataset's set of base vectors be $\mathcal{P}_o = \{\mathbf{p}_i\}^{n_o}_{i=1}$ and the set of query vectors be $\mathcal{Q} \subset \mathbb{R}^d$.  
We perform $s$ iterations of data updates, including insertions and deletions.  
The number of vectors in the index after each update is maintained at $n (\leq n_o)$, regardless of $s$.  
A full search is conducted on this index with the query set $\mathcal{Q}$ to obtain the ground-truth set $\mathcal{G}_s \subset \{1,2,\dots,n\}$ for each $s$, where $|\mathcal{G}_s| = |\mathcal{Q}|$.  
Data deletion and insertion are performed iteratively with an equal number of data points.
We define an insertion algorithm, \texttt{ADD}$: (\mathcal{P}_{\rm{add}},\mathcal{P},\mathcal{N}) \mapsto (\mathcal{P'},\mathcal{N'})$, which takes as input the set of vectors to be added $\mathcal{P}_{\rm{add}} \subset \mathbb{R}^d$, the existing node set $\mathcal{P} \subset \mathbb{R}^d$, and the neighborhood set $\mathcal{N}$, and outputs the updated node set $\mathcal{P'}$ and the updated neighborhood set $\mathcal{N'}$.  
Let the batch size for each insertion and deletion operation be $b \in \mathbb{N}$.  
At step $s=0$, the index is constructed with the first $n (\le n_o)$ base vectors as follows.  
\begin{equation}
    \label{eq:construct-k=0}
    \begin{split}
        \mathcal{P} &\leftarrow \{\mathbf{p}_i\}^{n}_{i=1} \\
        \mathcal{N} &\leftarrow \texttt{CONSTRUCT} (\mathcal{P})
    \end{split}
\end{equation}
For trials where $s \geq 1$, data deletion is first performed according to \eq{\ref{eq:delete-k>=1}}.  
Specifically, a set of $b$ consecutive integers is prepared as the deletion index set $\mathcal{D}$, and the vectors corresponding to these indices are removed.  
\begin{equation}
    \label{eq:delete-k>=1}
    \begin{split}
        \mathcal{D} &\leftarrow \{1+(s-1) b,\dots,sb\}\\
        (\mathcal{P},\mathcal{N}) &\leftarrow \texttt{DELETE} (\mathcal{D},\mathcal{P},\mathcal{N})
    \end{split}
\end{equation}
Next, $b$ data points are inserted.  
Specifically, a set of $b$ consecutive vectors, $\mathcal{P}_{\rm{add}}$, is prepared and added to the index, as seen below.  
\begin{equation}
    \label{eq:add-k>=1}
    \begin{split}
        \mathcal{P}_{\rm{add}} &\leftarrow \{\mathbf{p}_i \mid i \in \{1+n+(s-1) b,\dots,n+sb \} \} \\
        (\mathcal{P},\mathcal{N}) &\leftarrow \texttt{ADD} (\mathcal{P}_{\rm{add}},\mathcal{P},\mathcal{N})
    \end{split}
\end{equation}

The above experimental setup implies that when $b$ is small, frequent updates with a small number of data points occur.  
Conversely, when $b$ is large, a large amount of data is updated in a few iterations.  
The inserted data points are always new.  
All experiments were conducted on a single thread using an Intel(R) Core(TM) i7-13700H@2.4GHz processor with 32GB RAM, running Ubuntu 22.04.5.

\section{Mathematical Representation of Evaluation Metrics}  
The 1-Recall@$k$, which represents search accuracy, is defined as follows. Let \( n_q \) be the number of queries, and for a given query \( \mathbf{q}_i \in \mathbb{R}^d \), let \( g_i \in \{1,2,\dots,n\} \) denote the ground-truth nearest neighbor. Additionally, let \( \mathcal{\hat{R}}_i \subset \{1,2,\dots,n\} \) with \( |\mathcal{\hat{R}}_i| = k \) represent the approximate \( k \)-nearest neighbors obtained through ANNS. Defining \( f(\cdot) \) as a function that returns 1 if the condition is true and 0 otherwise, 1-Recall@$k$ is expressed as shown in Equation~\ref{eq:recall_at_k}.

\begin{equation}
\label{eq:recall_at_k}
    {\rm{1}\text{-}\rm{Recall}}@k
    = \frac{1}{n_q} \sum_{i=1}^{n_q} 
      f\bigl(g_i \in \mathcal{\hat{R}}_{i}\bigr),
\end{equation}
In this study, we set \( k = 10 \) and use 1-Recall@10 to evaluate search accuracy.
A higher recall indicates better search accuracy.  

To evaluate the query processing speed, we use Queries Per Second (QPS). When processing \( n_q \) queries in \( t \) seconds, QPS is defined as shown in Equation~\ref{eq:qps}:  
\begin{equation}
\label{eq:qps}
    {\rm{QPS}} 
    = \frac{n_q}{t} [1/{\rm{s}}]
\end{equation}
A higher QPS value indicates faster query processing. We measure QPS-search to evaluate search speed, QPS-add to evaluate data insertion speed, and QPS-delete to evaluate data deletion speed.  
Additionally, we adopt the QPS-Recall curve as an evaluation metric, where the horizontal axis represents 1-Recall@10, and the vertical axis represents QPS-search. This curve is obtained by varying the search parameters of HNSW. A curve positioned toward the upper right of the graph indicates higher search performance.

\section{All Experimental Results} 
\label{sec:all-experimental-results}
The experimental results for SIFT1M~\cite{sift1m} are shown in \fig{\ref{fig:sift1m}}, those for GIST1M~\cite{sift1m} are presented in \fig{\ref{fig:gist1m}}, those for DEEP1M~\cite{deep1m} are presented in \fig{\ref{fig:deep1m}}, and those for Glove100Angular~\cite{glove} are presented in \fig{\ref{fig:glove100}}. The results for SIFT1B~\cite{sift1b} with a batch size of \( b = 10^5 \) are shown in \fig{\ref{fig:sift1b2m-100000}}, while those with \( b = 10^3 \) are given in \fig{\ref{fig:sift1b2m-1000}}.  
In the following sections, we discuss the experimental results for each evaluation metric.

\subsection{QPS-Recall}\label{subsec:qps-recall}
From \fig{\ref{fig:sift1m-qps-recall}}, \fig{\ref{fig:gist1m-qps-recall}}, \fig{\ref{fig:sift1b2m-qps-recall-100000}}, \fig{\ref{fig:sift1b2m-qps-recall-1000}}, \fig{\ref{fig:deep1m-qps-recall}} and \fig{\ref{fig:glove100-qps-recall}}, it is evident that rebuilding maintains search performance even after repeated insertions and deletions.  
In contrast, logical deletion significantly degrades search performance across all datasets as insertions and deletions are repeated. The plotted points in each graph indicate that both search accuracy and search speed deteriorate in this case.  
Furthermore, the search performance of physical deletion is slightly lower than that of rebuilding across all datasets.

\subsection{1-Recall@10}\label{subsec:1recall@10}
From \fig{\ref{fig:sift1m-recall}}, \fig{\ref{fig:gist1m-recall}}, \fig{\ref{fig:sift1b2m-recall-100000-supp}}, \fig{\ref{fig:sift1b2m-recall-1000}}, \fig{\ref{fig:deep1m-recall}} and \fig{\ref{fig:glove100-recall}}, it can be observed that search accuracy in logical deletion decreases as insertions and deletions are repeated.  
Additionally, the accuracy of rebuilding is the highest, followed by physical deletion, which exhibits lower accuracy than rebuilding.  
Furthermore, \fig{\ref{fig:sift1b2m-recall-100000-supp}} and \fig{\ref{fig:sift1b2m-recall-1000}} show that in physical deletion, search accuracy stabilizes after a certain number of insertion and deletion steps. This indicates that the structural properties of the graph become stable after a sufficient number of operations.  
Moreover, a larger batch size results in a higher converged accuracy. This suggests that when insertions and deletions are performed repeatedly, a larger batch size facilitates better recovery of the graph structure during the insertion process.

\subsection{Memory Usage}\label{subsec:memory-usage}  
From \fig{\ref{fig:sift1m-memory}}, \fig{\ref{fig:gist1m-memory}}, \fig{\ref{fig:sift1b2m-memory-100000}}, \fig{\ref{fig:sift1b2m-memory-1000}}, \fig{\ref{fig:deep1m-memory}} and \fig{\ref{fig:glove100-memory}}, it is evident that memory consumption in logical deletion increases linearly with each step across all datasets. This indicates that in logical deletion, the deleted data remains in memory.  
In contrast, memory usage remains unchanged for both rebuilding and physical deletion. This confirms that these methods effectively reclaim memory space when data is deleted.  

\subsection{QPS-add}\label{subsec:qps-add}
From \fig{\ref{fig:sift1m-qps-add}}, \fig{\ref{fig:gist1m-qps-add}}, \fig{\ref{fig:sift1b2m-qps-add-100000}},  \fig{\ref{fig:deep1m-qps-add}} and \fig{\ref{fig:glove100-qps-add}}, it can be observed that when data is inserted and deleted in batches of \( b = 10^5 \), the data insertion speed remains unchanged.  
However, as shown in \fig{\ref{fig:sift1b2m-qps-add-1000}}, when the batch size is reduced to \( b = 10^3 \), the data insertion speed in logical deletion exhibits significant variations at each step.  

Additionally, when data insertion and deletion are performed in batches of \( b = 10^5 \), physical deletion exhibits the highest data insertion speed. This is likely because repeated physical deletions gradually make the graph sparser, thereby reducing the number of distance calculations required during data insertion.

\subsection{QPS-delete}\label{subsec:qps-delete}
\begin{figure*}[tb]
    \centering
    \begin{subfigure}{0.32\textwidth}
        \centering
        \includegraphics[width=\textwidth]{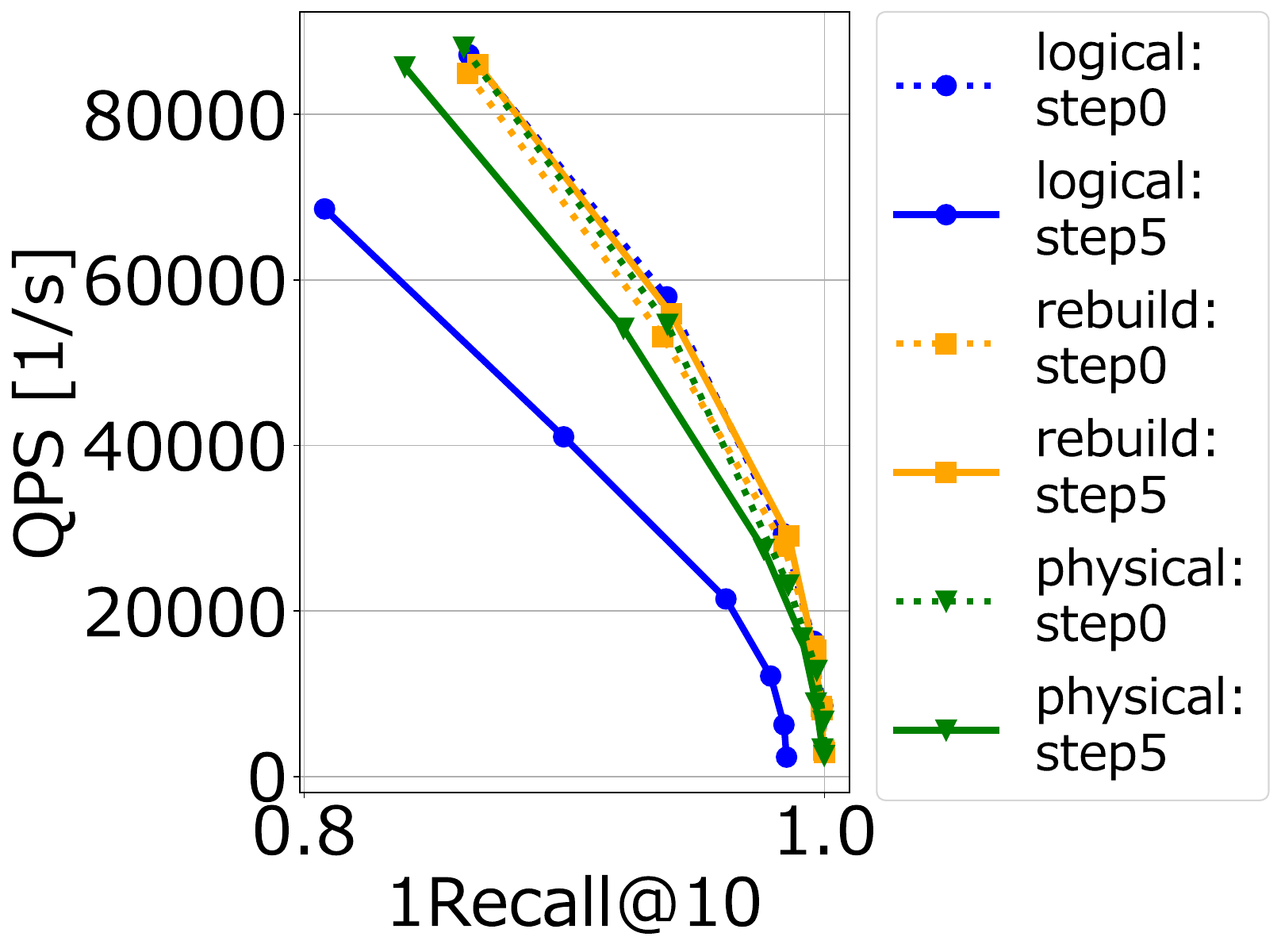}
        \caption{QPS-Recall}
        \label{fig:sift1m-qps-recall}
    \end{subfigure}
    \hfill
    \begin{subfigure}{0.32\textwidth}
        \centering
        \includegraphics[width=\textwidth]{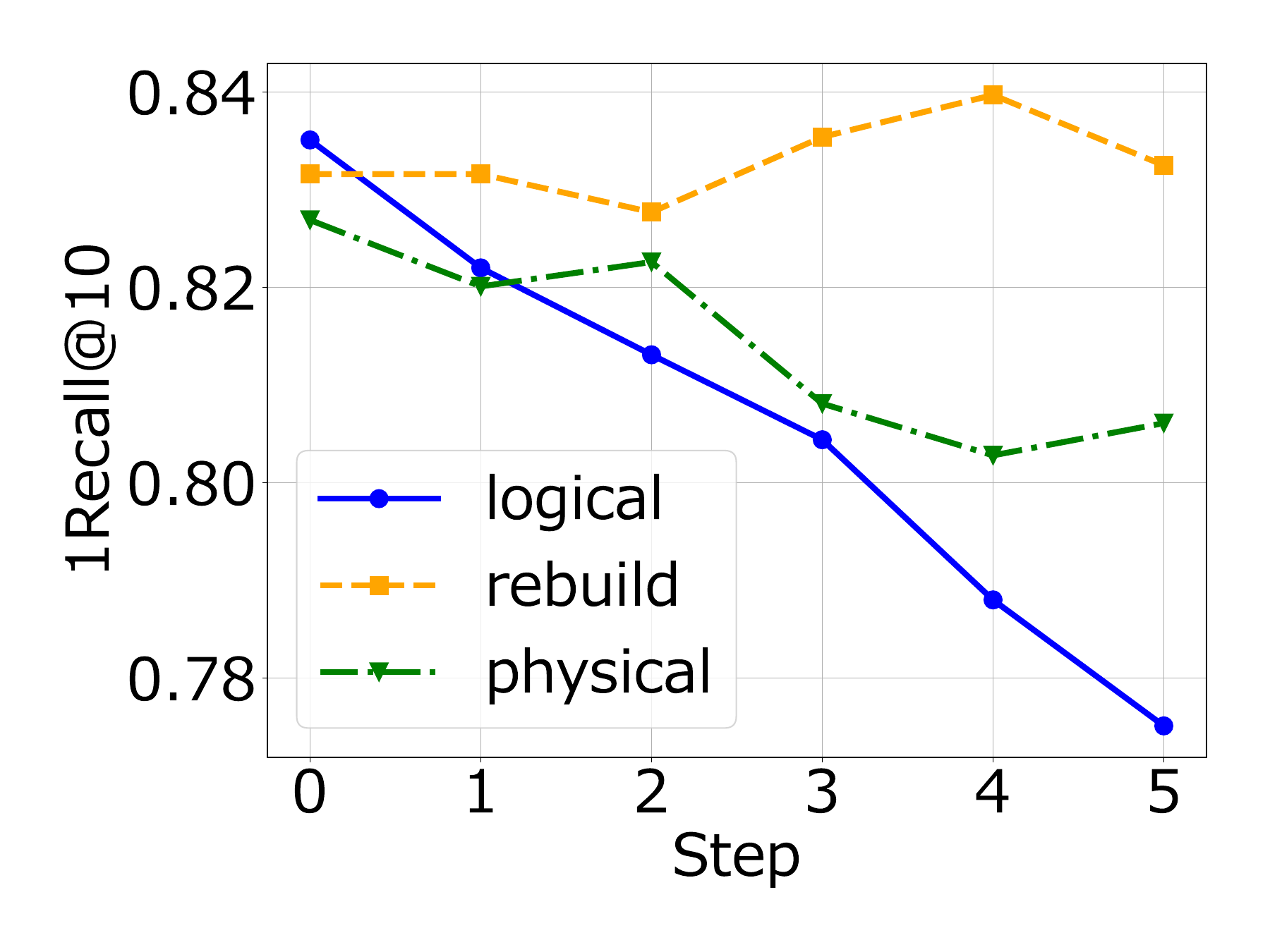}
        \caption{1-Recall@10}
        \label{fig:sift1m-recall}
    \end{subfigure}
    \hfill
    \begin{subfigure}{0.32\textwidth}
        \centering
        \includegraphics[width=\textwidth]{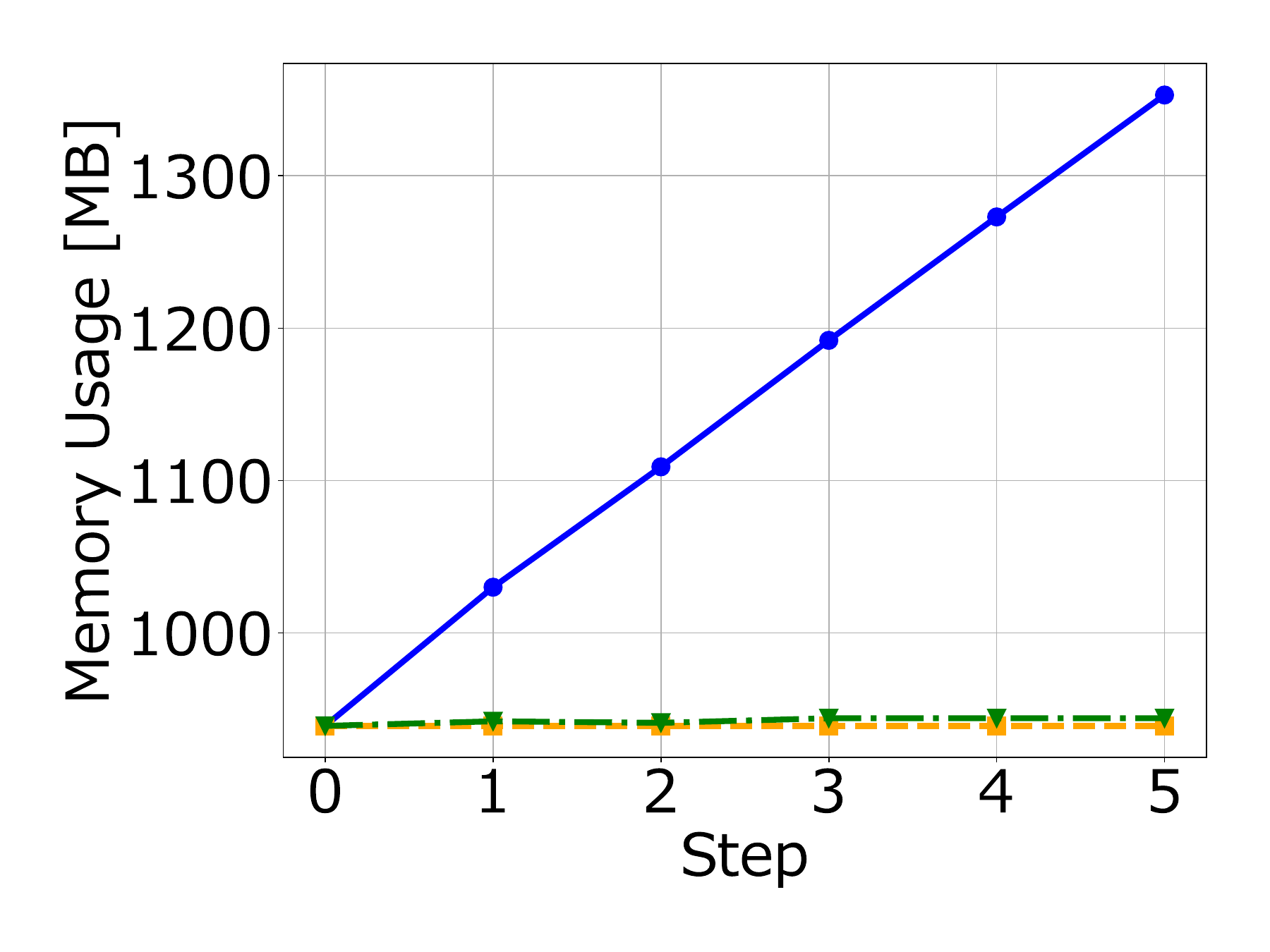}
        \caption{Memory Usage}
        \label{fig:sift1m-memory}
    \end{subfigure}
    \hfill
    \begin{subfigure}{0.32\textwidth}
        \centering
        \includegraphics[width=\textwidth]{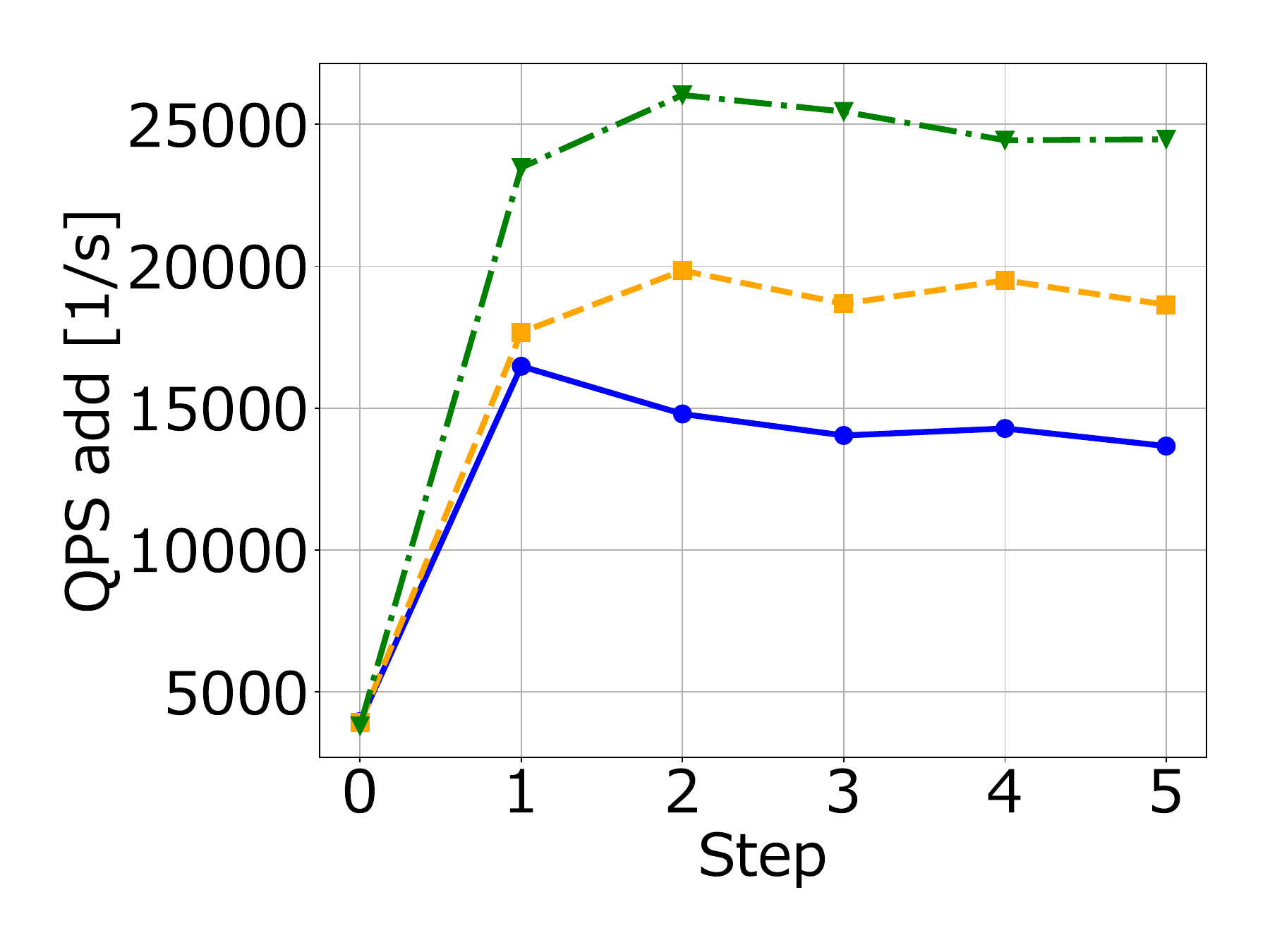}
        \caption{QPS-add}
        \label{fig:sift1m-qps-add}
    \end{subfigure}
    \hfill
    \begin{subfigure}{0.32\textwidth}
        \centering
        \includegraphics[width=\textwidth]{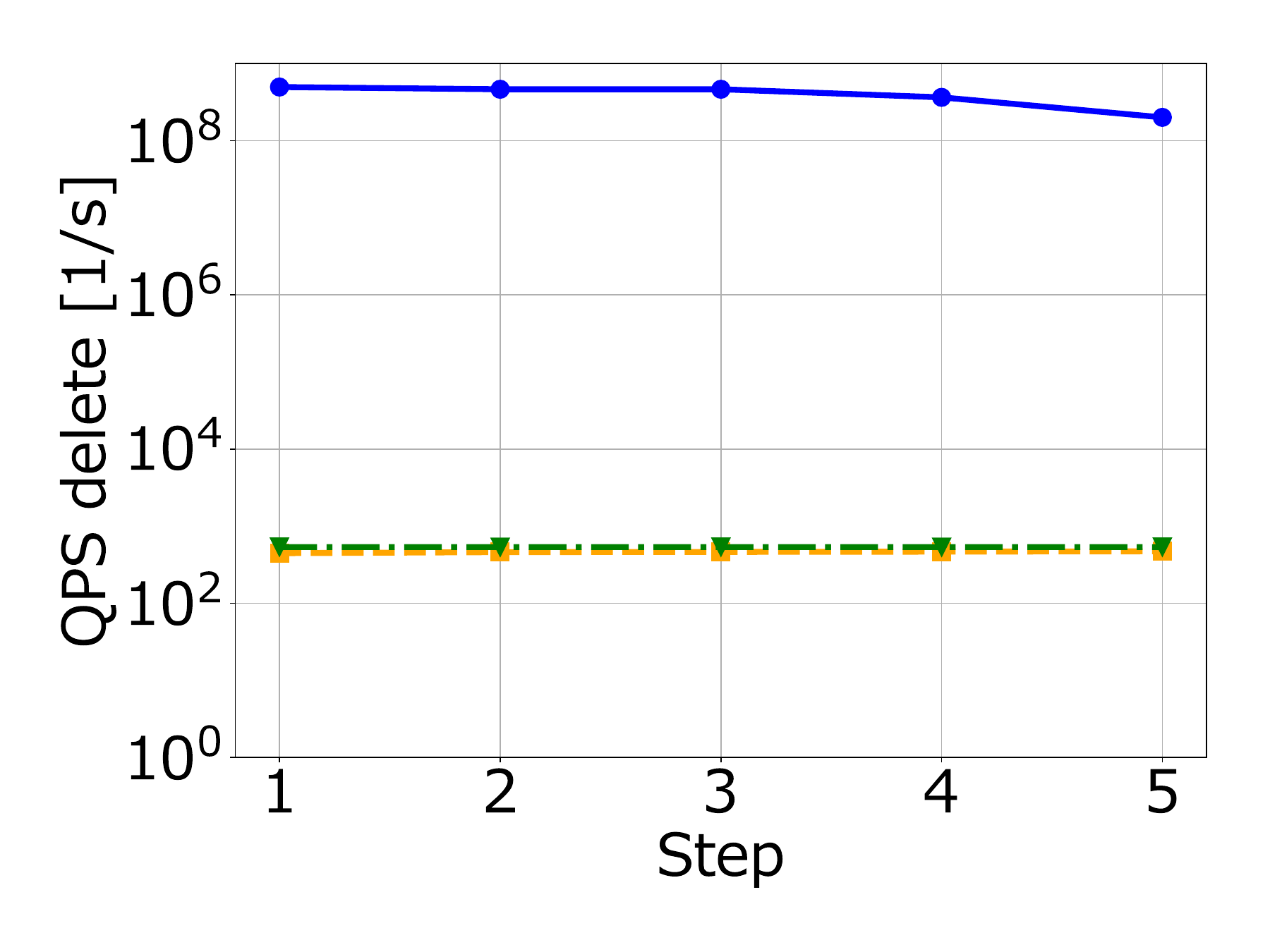}
        \caption{QPS-delete}
        \label{fig:sift1m-qps-delete}
    \end{subfigure}
    \hfill
    \begin{subfigure}{0.32\textwidth}
        \centering
        \includegraphics[width=\textwidth]{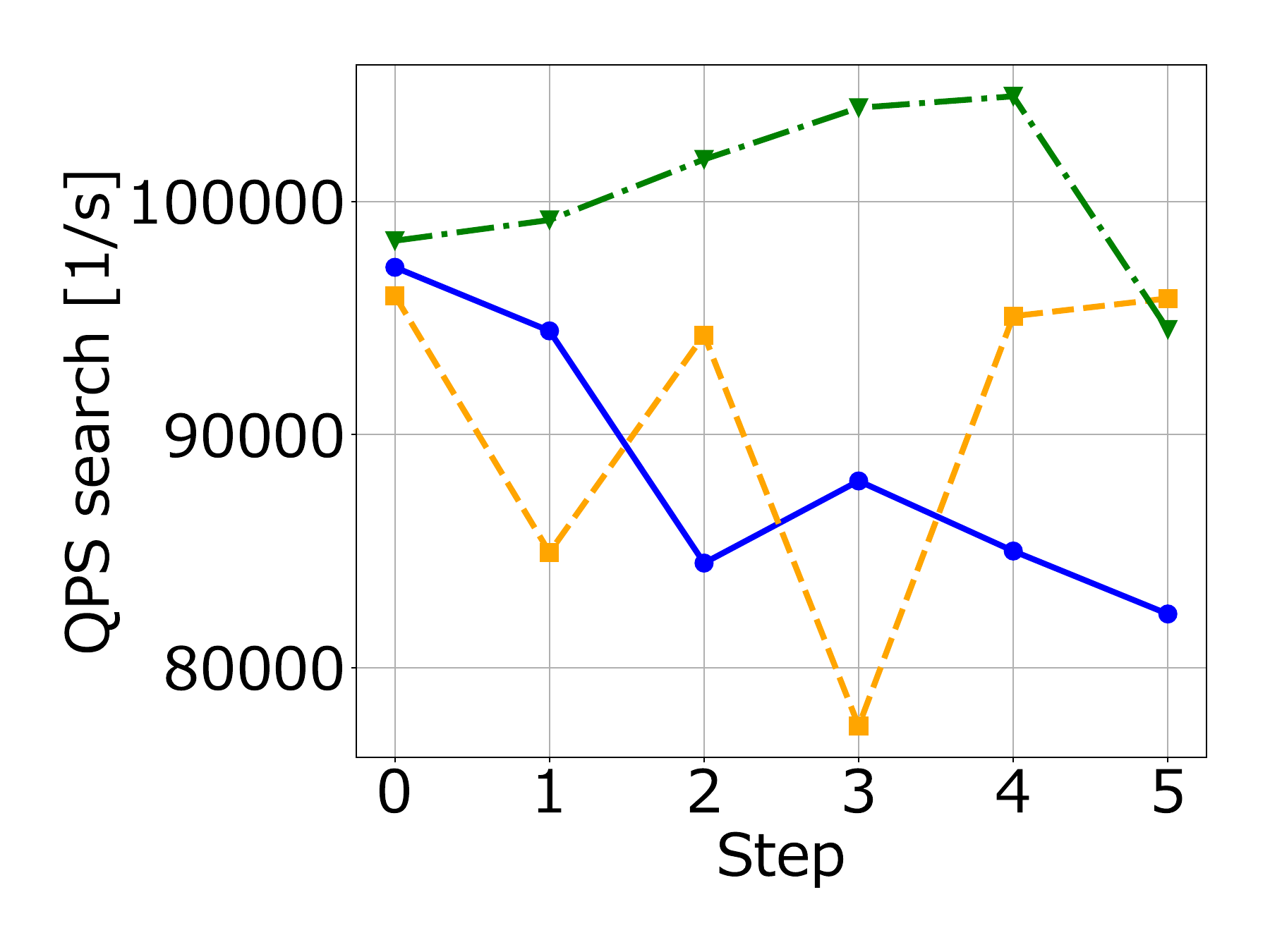}
        \caption{QPS-search}
        \label{fig:sift1m-qps-search}
    \end{subfigure}
    \hfill
    \caption{Performance comparison of the three deletion methods at each step on SIFT1M.}
    \label{fig:sift1m}
\end{figure*}

\begin{figure*}[tb]
    \centering
    \begin{subfigure}{0.32\textwidth}
        \centering
        \includegraphics[width=\textwidth]{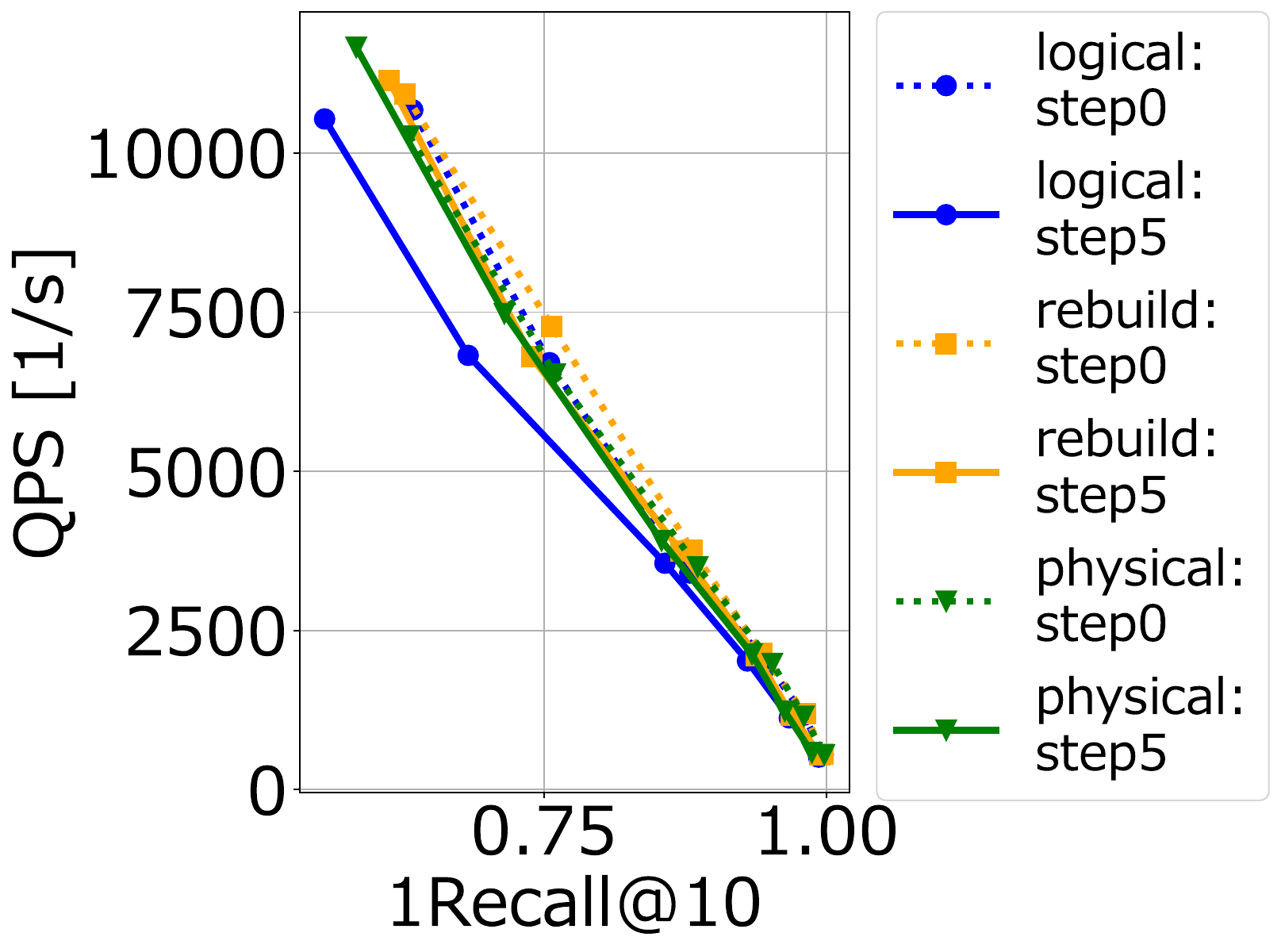}
        \caption{QPS-Recall}
        \label{fig:gist1m-qps-recall}
    \end{subfigure}
    \hfill
    \begin{subfigure}{0.32\textwidth}
        \centering
        \includegraphics[width=\textwidth]{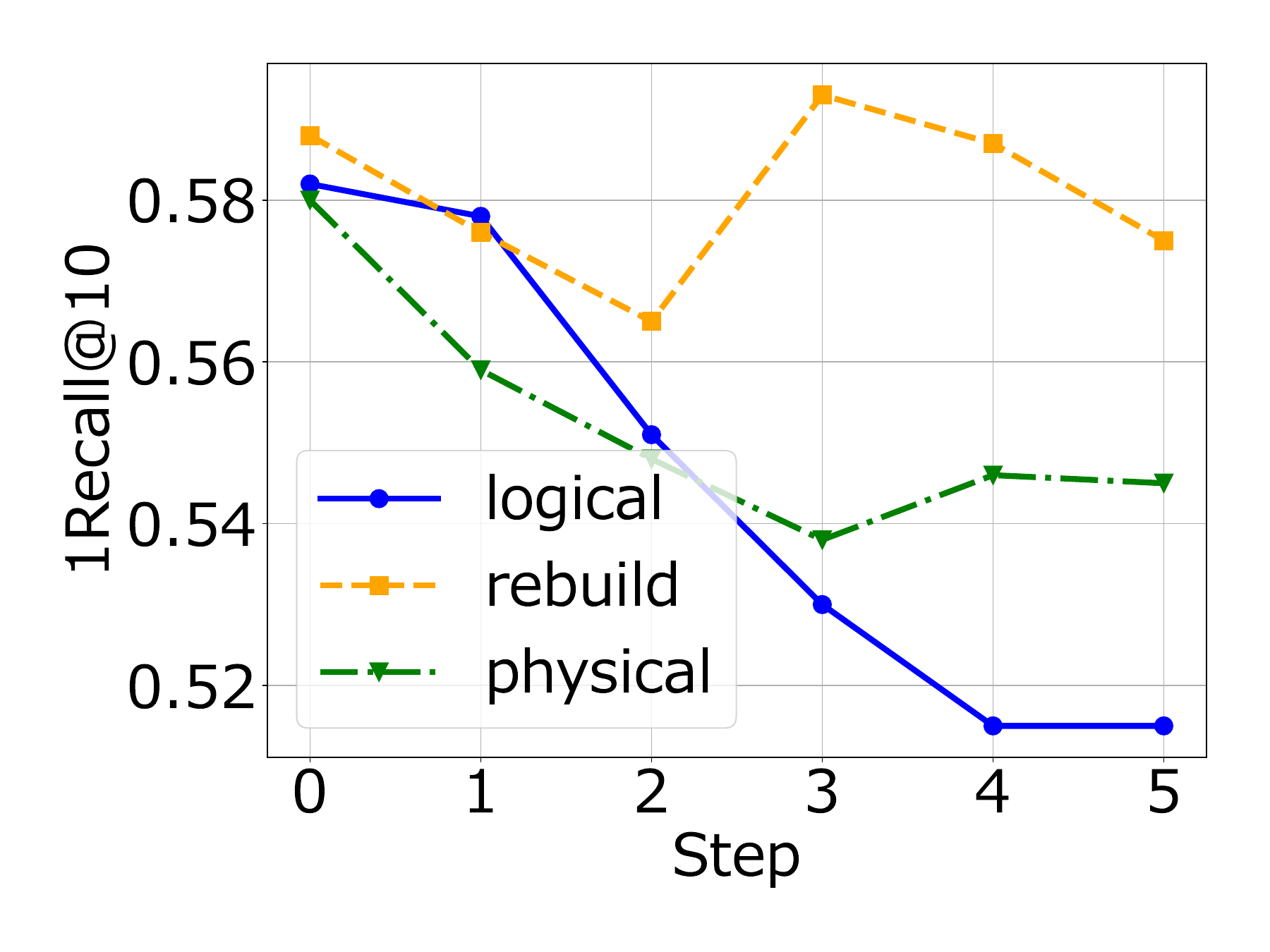}
        \caption{1-Recall@10}
        \label{fig:gist1m-recall}
    \end{subfigure}
    \hfill
    \begin{subfigure}{0.32\textwidth}
        \centering
        \includegraphics[width=\textwidth]{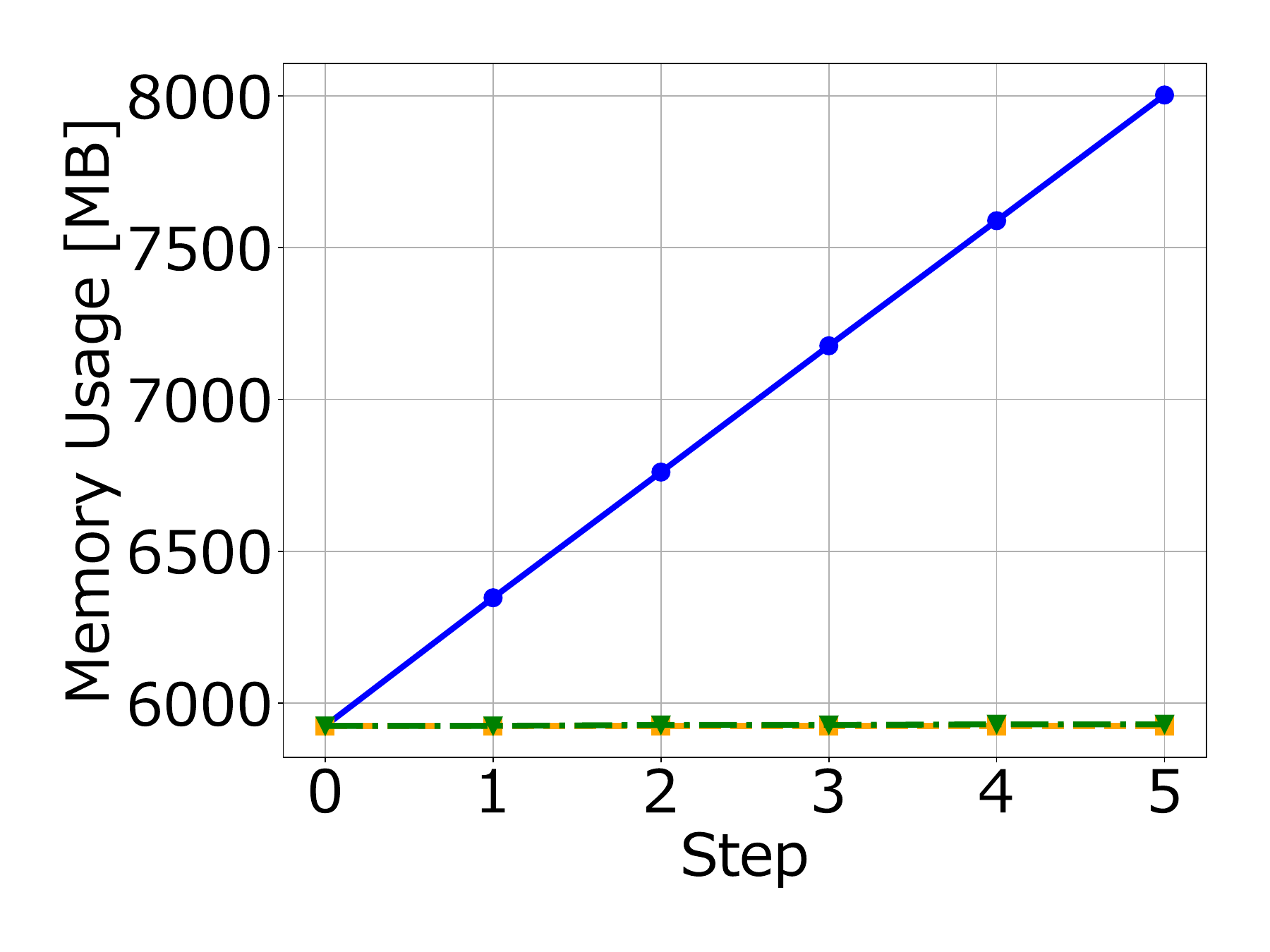}
        \caption{Memory Usage}
        \label{fig:gist1m-memory}
    \end{subfigure}
    \hfill
    \begin{subfigure}{0.32\textwidth}
        \centering
        \includegraphics[width=\textwidth]{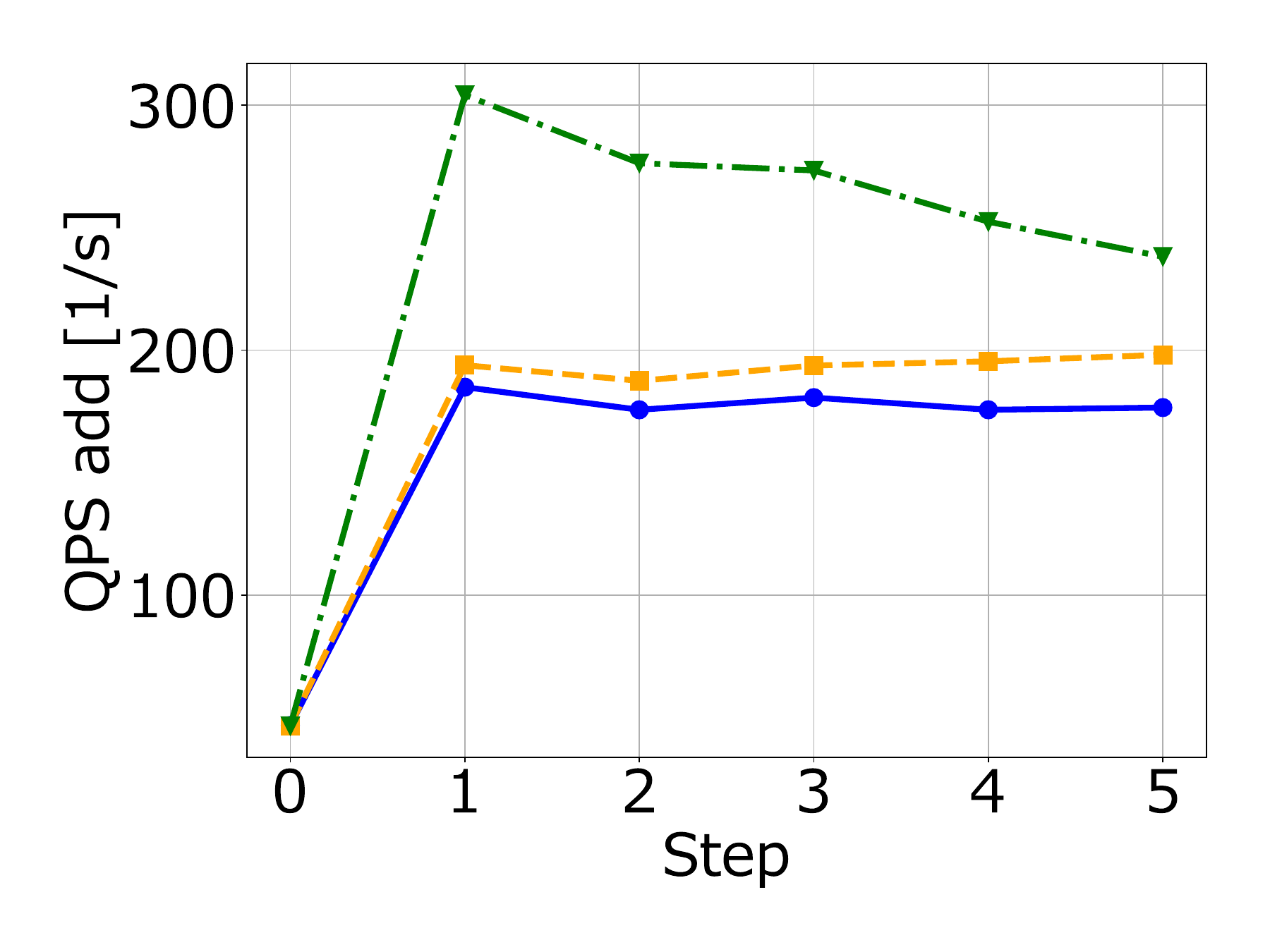}
        \caption{QPS-add}
        \label{fig:gist1m-qps-add}
    \end{subfigure}
    \hfill
    \begin{subfigure}{0.32\textwidth}
        \centering
        \includegraphics[width=\textwidth]{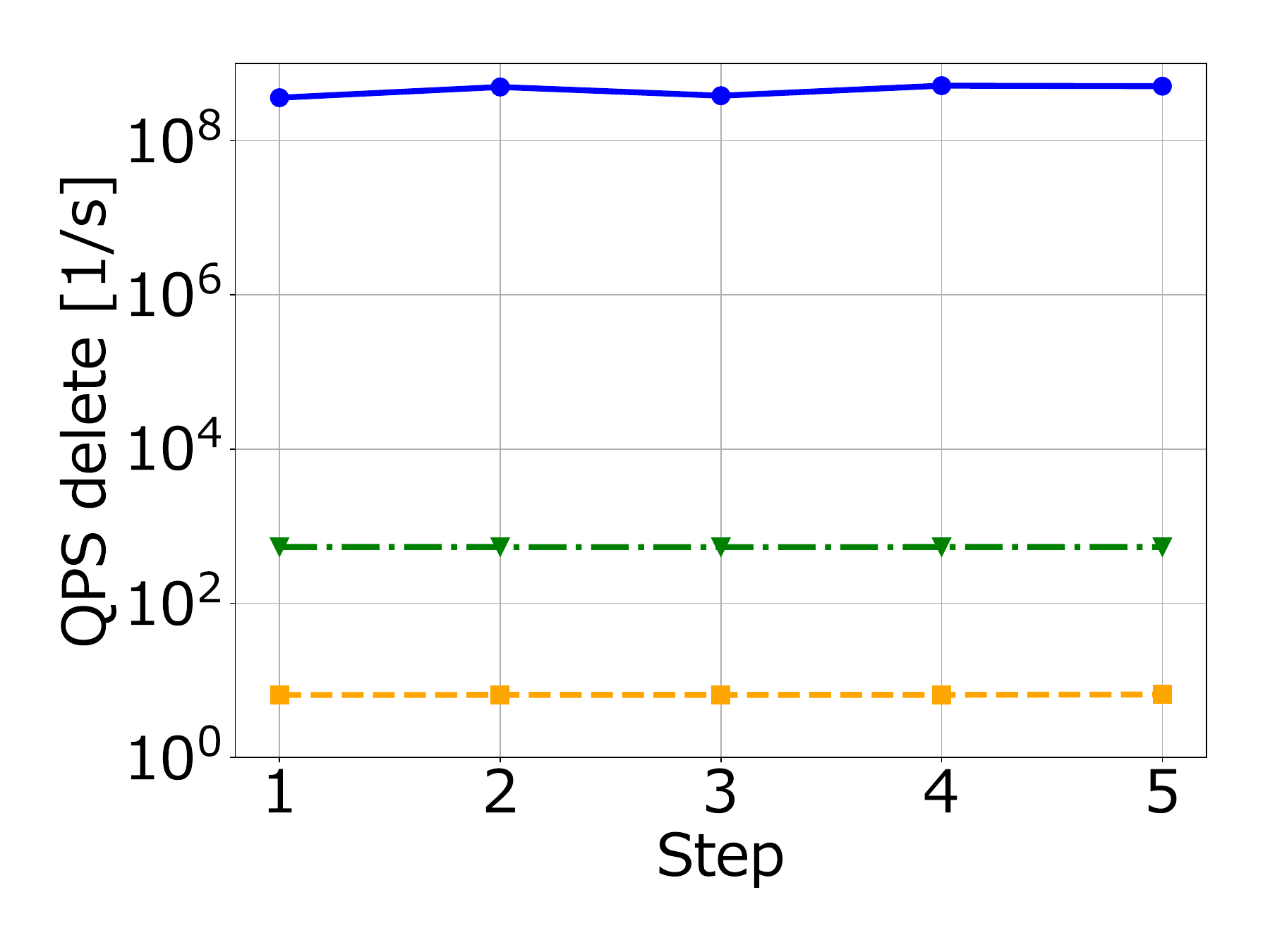}
        \caption{QPS-delete}
        \label{fig:gist1m-qps-delete}
    \end{subfigure}
    \hfill
    \begin{subfigure}{0.32\textwidth}
        \centering
        \includegraphics[width=\textwidth]{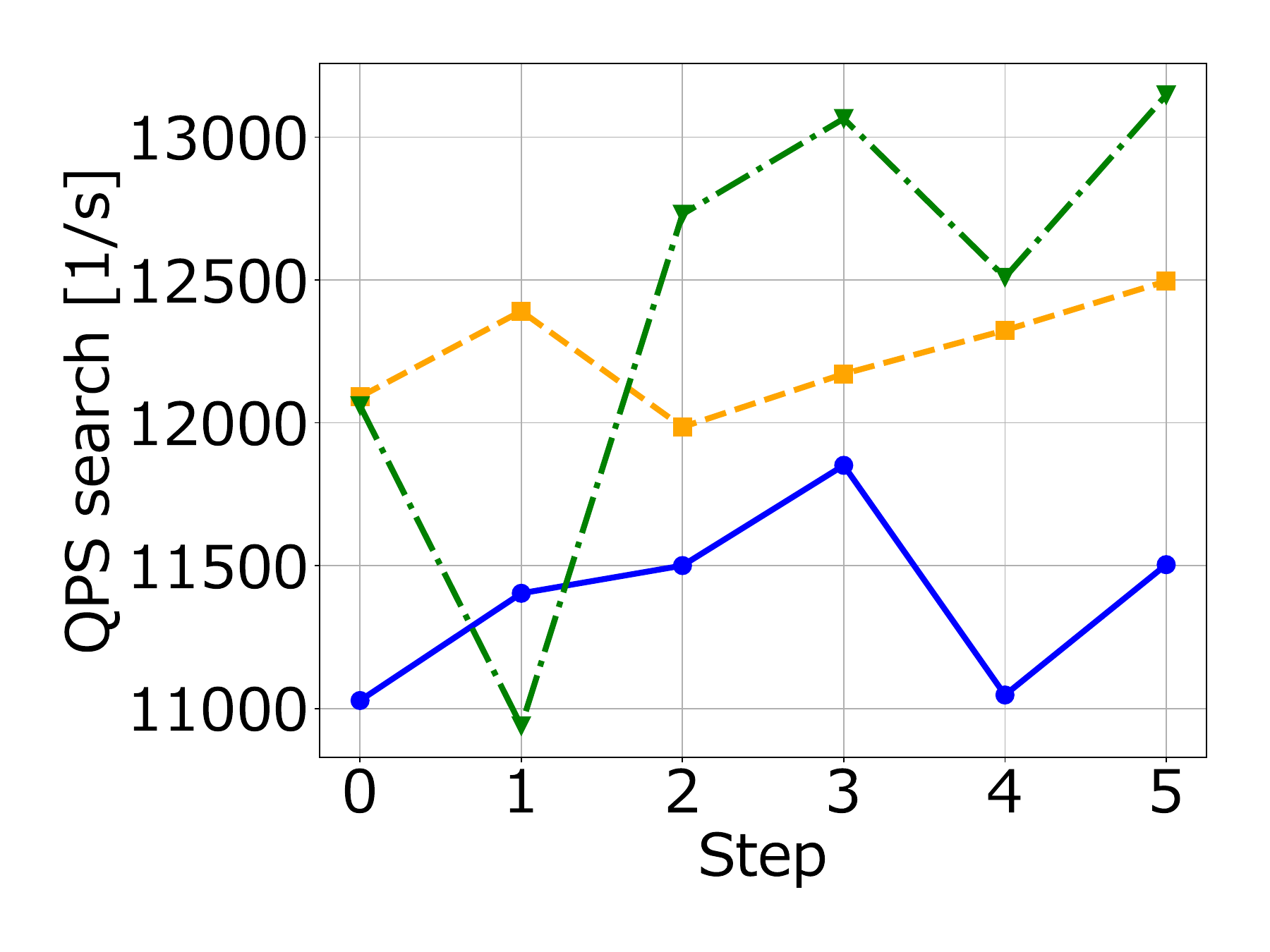}
        \caption{QPS-search}
        \label{fig:gist1m-qps-search}
    \end{subfigure}
    \hfill
    \caption{Performance comparison of the three deletion methods at each step on GIST1M.}
    \label{fig:gist1m}
\end{figure*}

\begin{figure*}[tb]
    \centering
    \begin{subfigure}{0.32\textwidth}
        \centering
        \includegraphics[width=\textwidth]{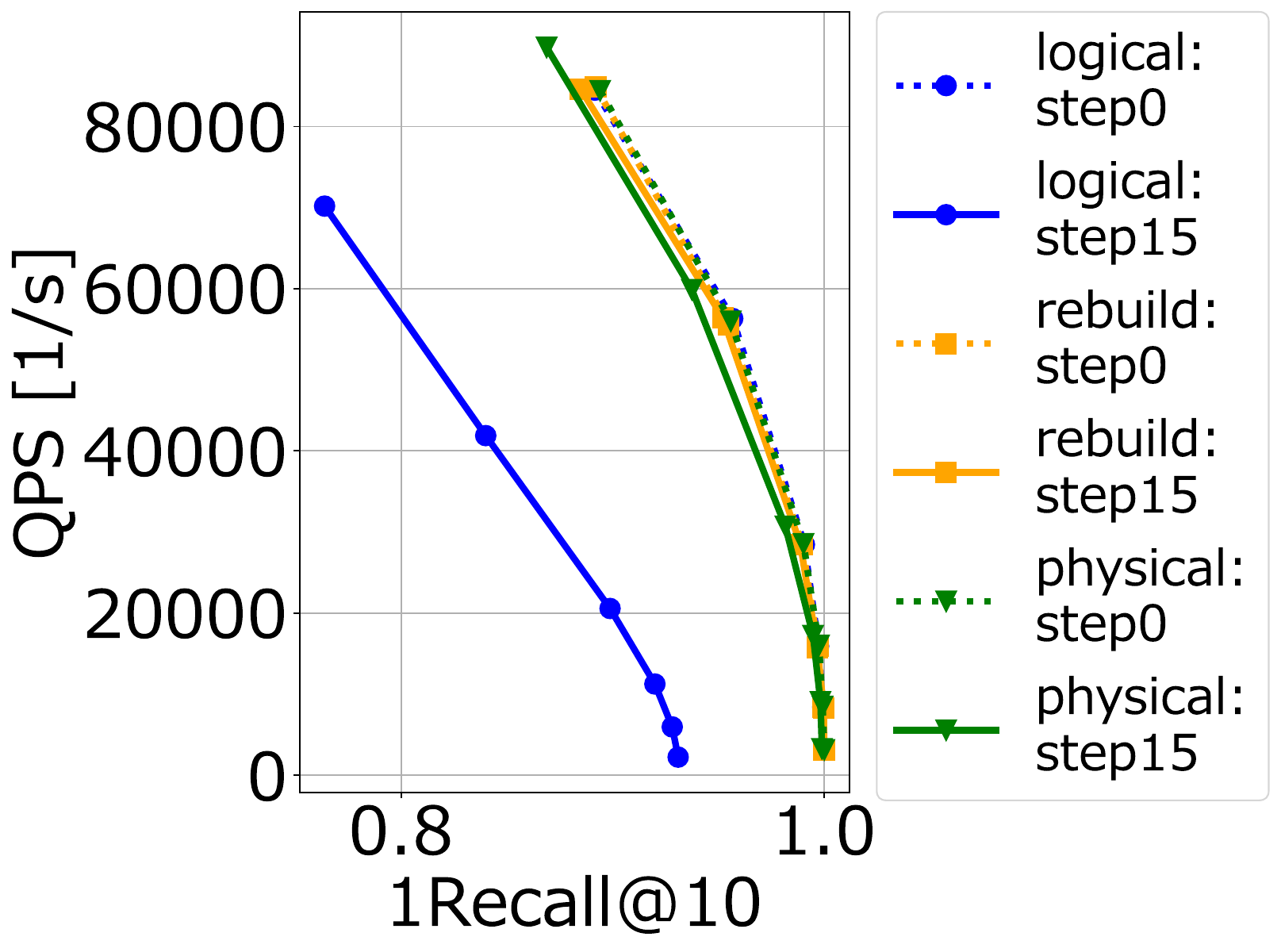}
        \caption{QPS-Recall}
        \label{fig:sift1b2m-qps-recall-100000}
    \end{subfigure}
    \hfill
    \begin{subfigure}{0.32\textwidth}
        \centering
        \includegraphics[width=\textwidth]{fig/experiment/sift1b_2m_1Recall_10_unit_100000.pdf}
        \caption{1-Recall@10}
        \label{fig:sift1b2m-recall-100000-supp}
    \end{subfigure}
    \hfill
    \begin{subfigure}{0.32\textwidth}
        \centering
        \includegraphics[width=\textwidth]{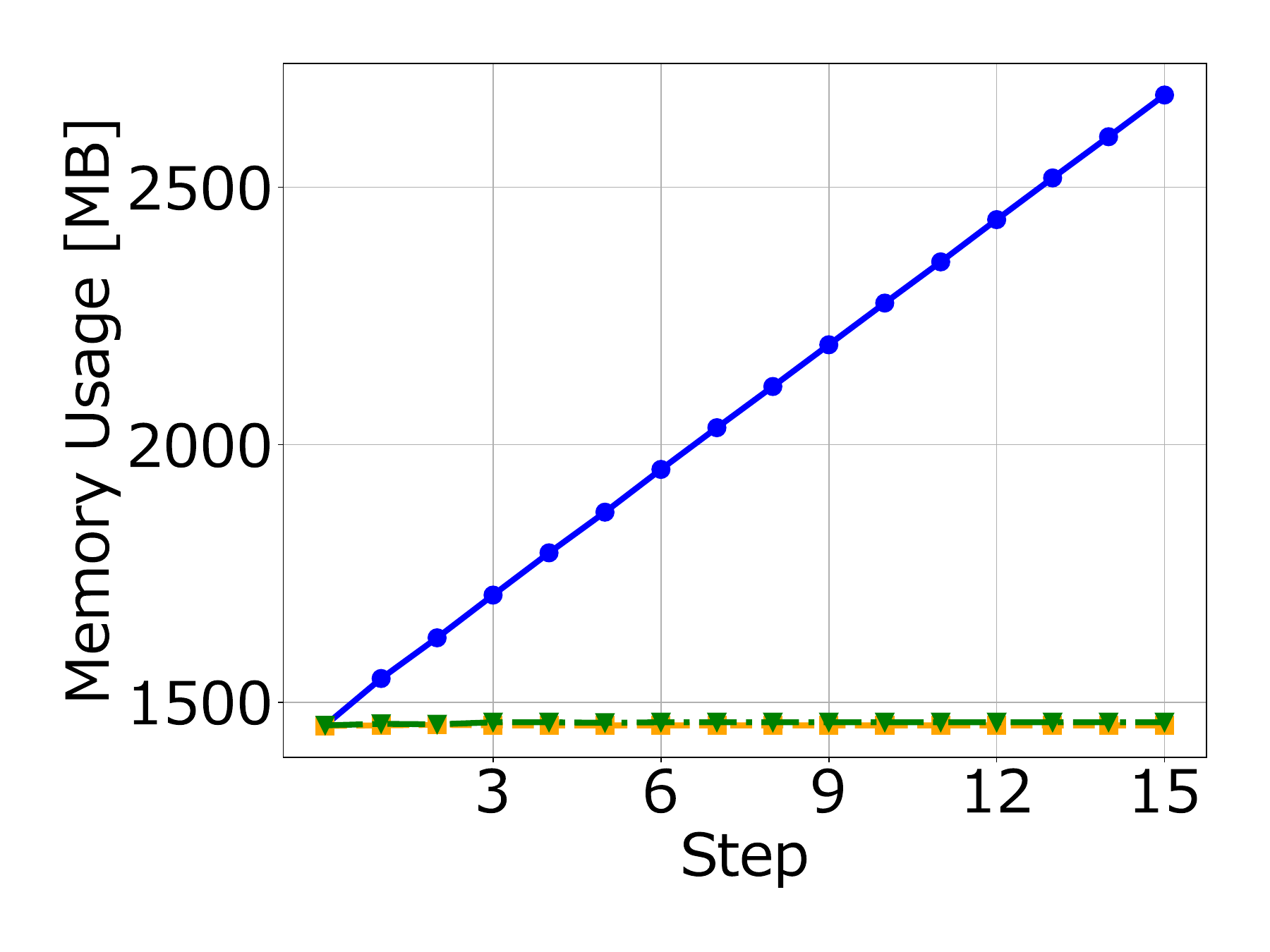}
        \caption{Memory Usage}
        \label{fig:sift1b2m-memory-100000}
    \end{subfigure}
    \hfill
    \begin{subfigure}{0.32\textwidth}
        \centering
        \includegraphics[width=\textwidth]{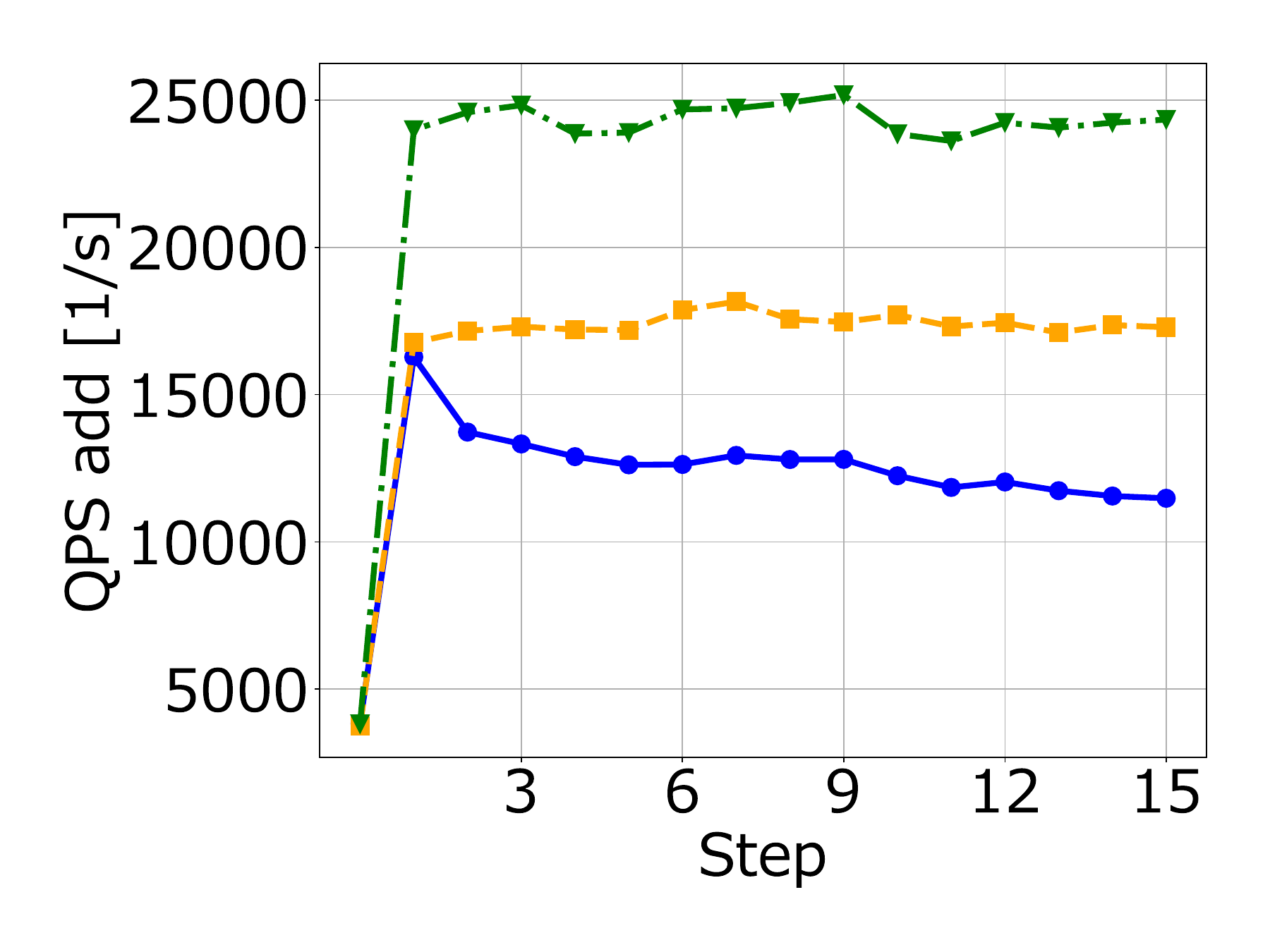}
        \caption{QPS-add}
        \label{fig:sift1b2m-qps-add-100000}
    \end{subfigure}
    \hfill
    \begin{subfigure}{0.32\textwidth}
        \centering
        \includegraphics[width=\textwidth]{fig/experiment/sift1b_2m_QPS_delete_unit_100000.pdf}
        \caption{QPS-delete}
        \label{fig:sift1b2m-qps-delete-100000-supp}
    \end{subfigure}
    \hfill
    \begin{subfigure}{0.32\textwidth}
        \centering
        \includegraphics[width=\textwidth]{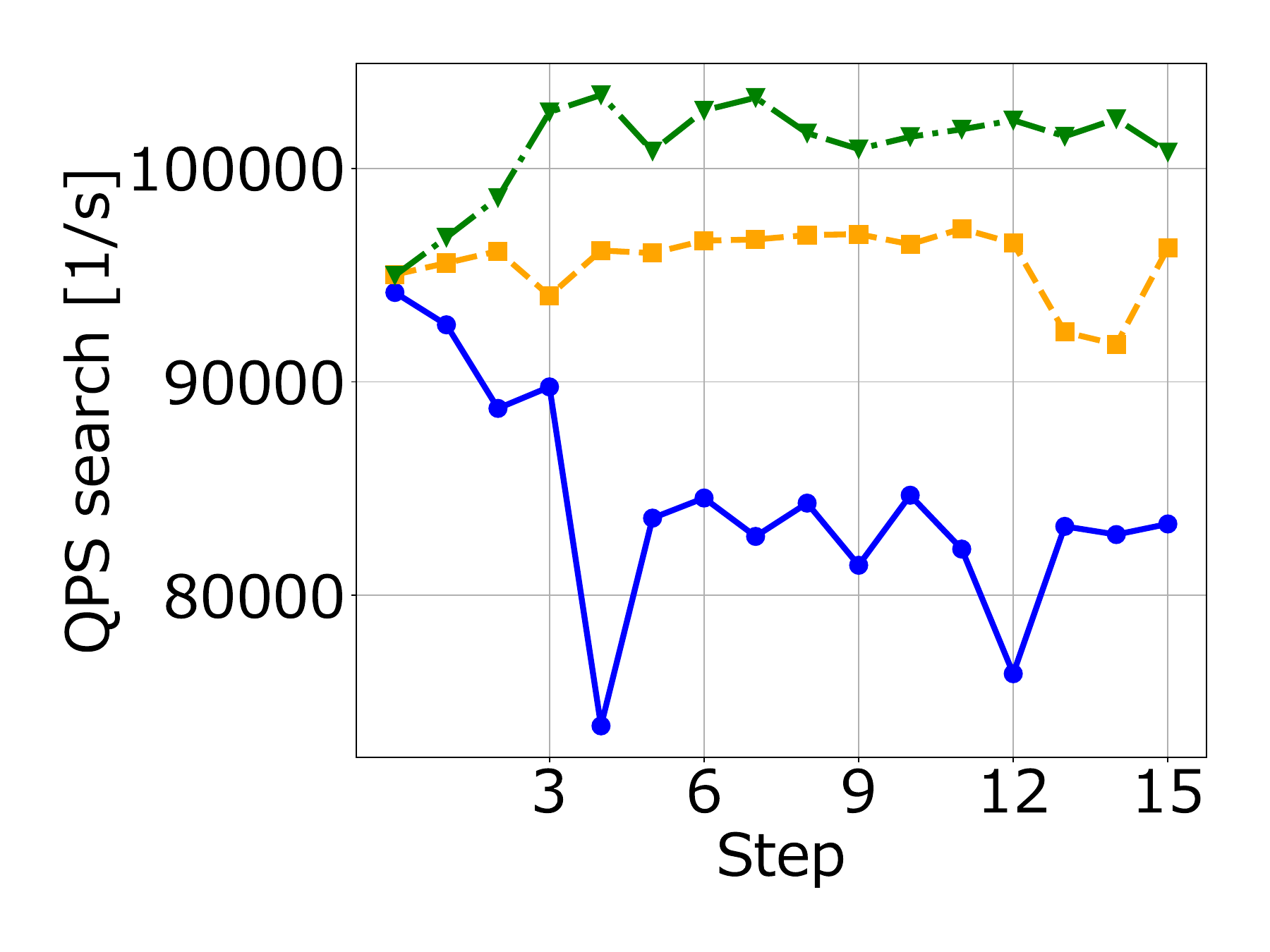}
        \caption{QPS-search}
        \label{fig:sift1b2m-qps-search-100000}
    \end{subfigure}
    \hfill
    \caption{Performance comparison of the three deletion methods at each step on SIFT1B with $b=10^5$.}
    \label{fig:sift1b2m-100000}
\end{figure*}

\begin{figure*}[tb]
    \centering
    \begin{subfigure}{0.32\textwidth}
        \centering
        \includegraphics[width=\textwidth]{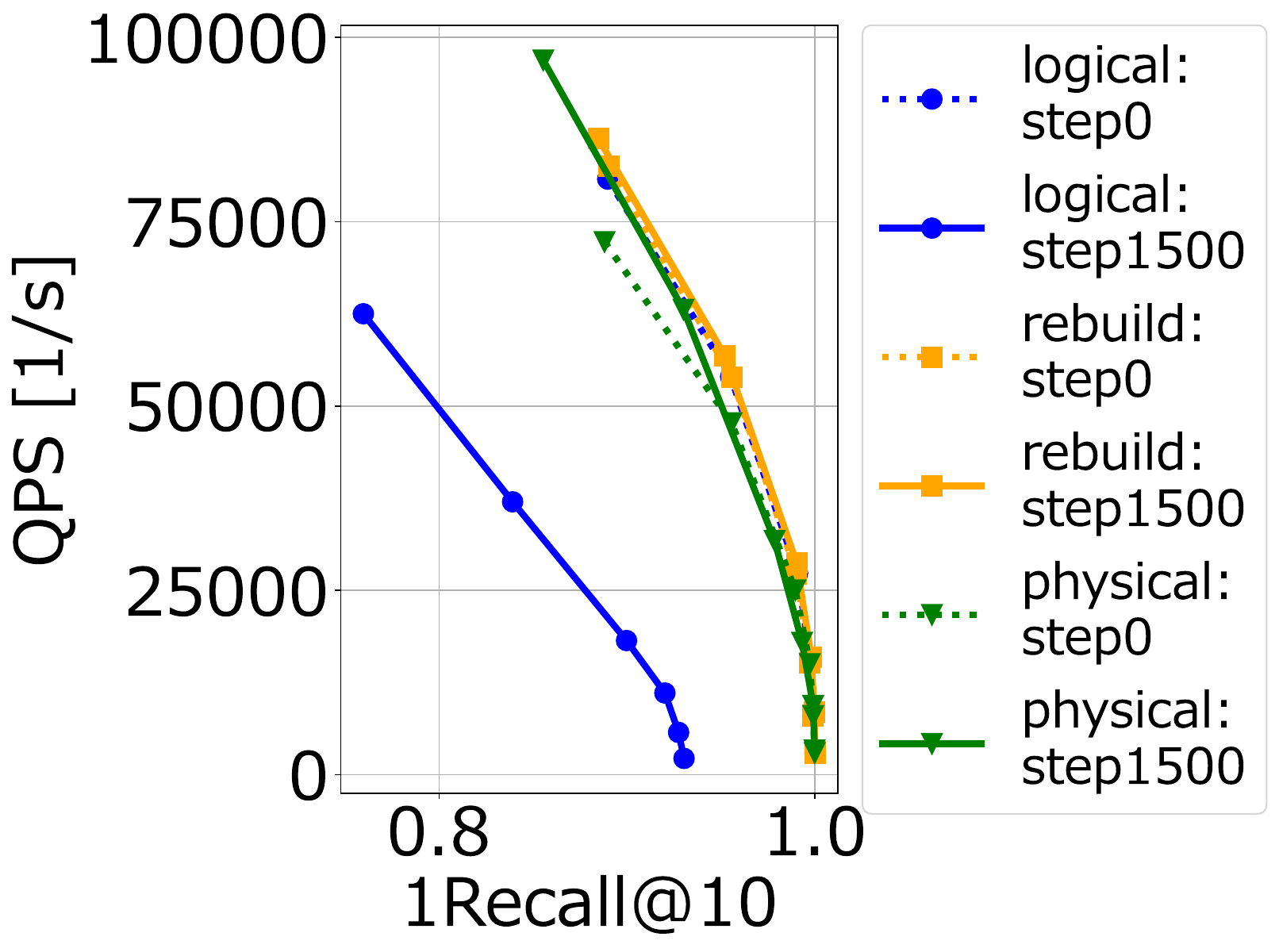}
        \caption{QPS-Recall}
        \label{fig:sift1b2m-qps-recall-1000}
    \end{subfigure}
    \hfill
    \begin{subfigure}{0.32\textwidth}
        \centering
        \includegraphics[width=\textwidth]{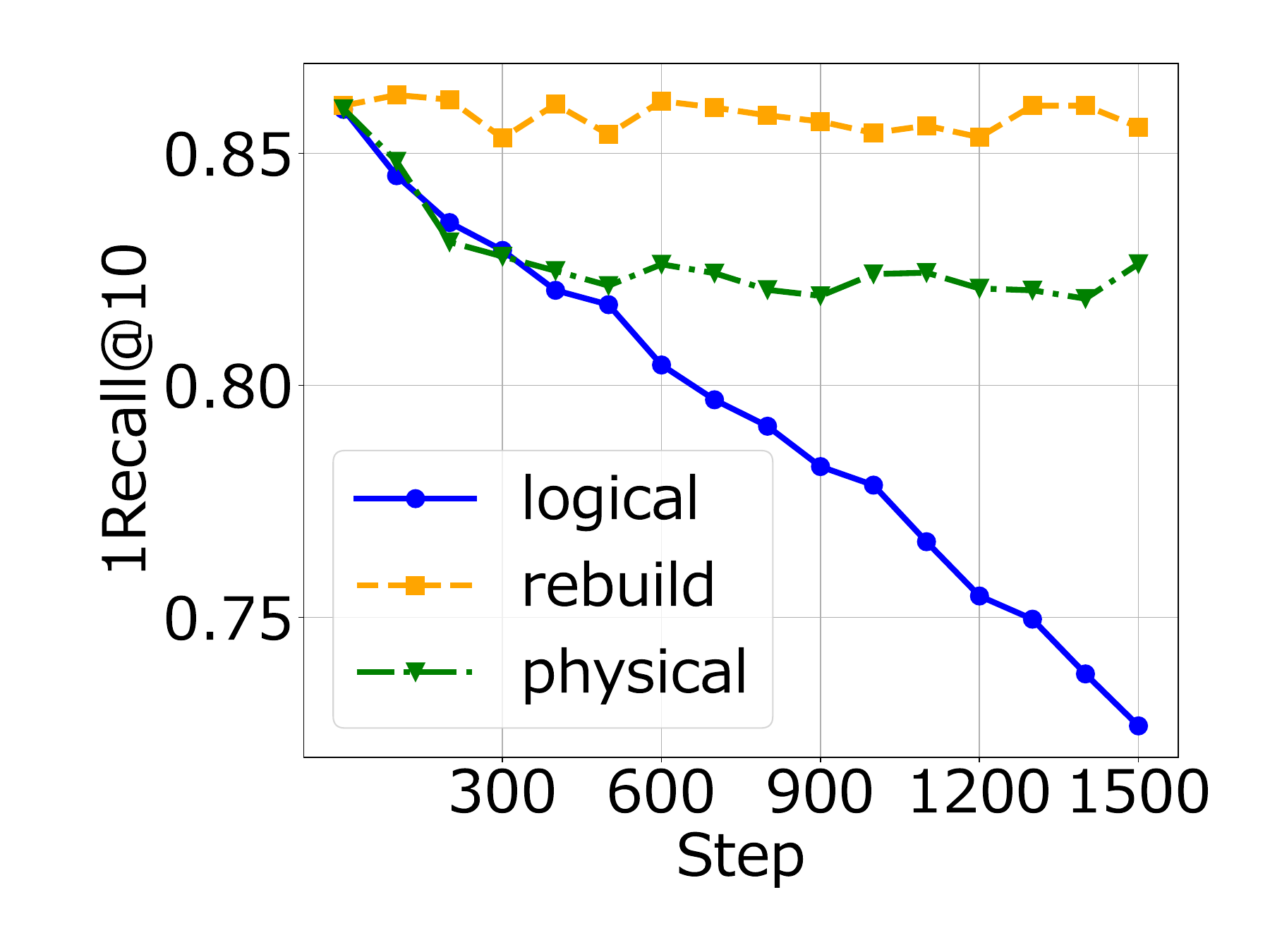}
        \caption{1-Recall@10}
        \label{fig:sift1b2m-recall-1000}
    \end{subfigure}
    \hfill
    \begin{subfigure}{0.32\textwidth}
        \centering
        \includegraphics[width=\textwidth]{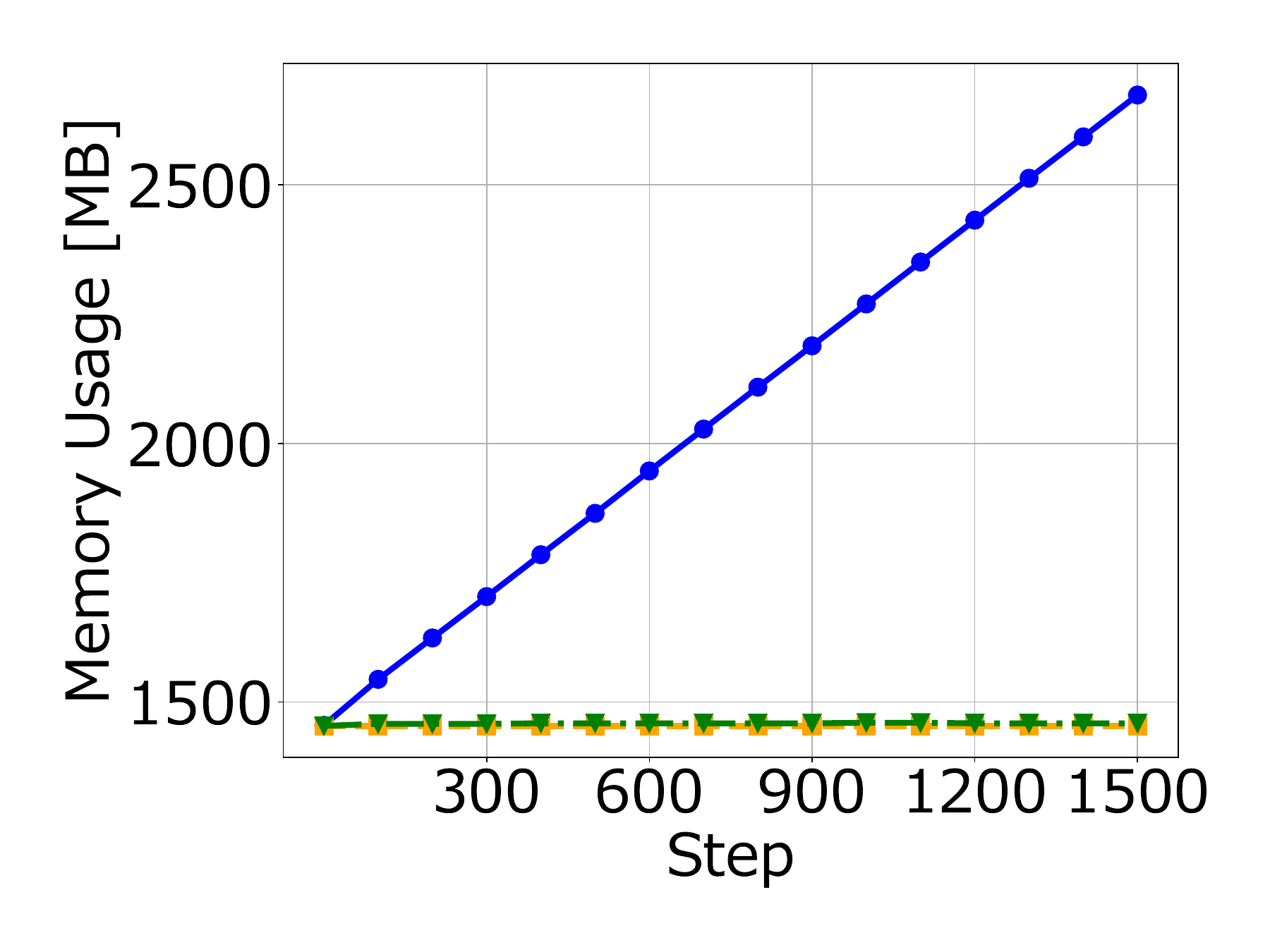}
        \caption{Memory Usage}
        \label{fig:sift1b2m-memory-1000}
    \end{subfigure}
    \hfill
    \begin{subfigure}{0.32\textwidth}
        \centering
        \includegraphics[width=\textwidth]{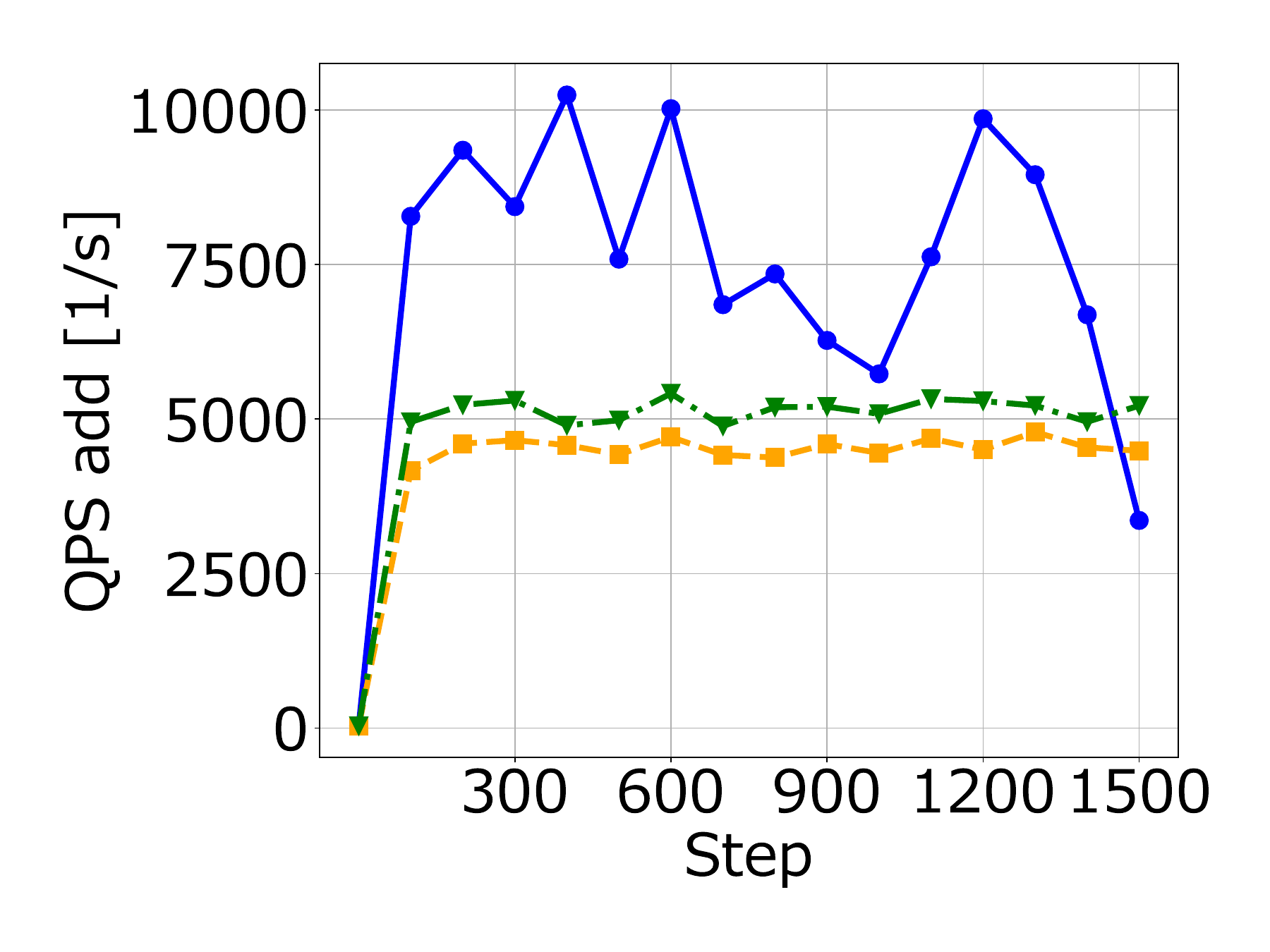}
        \caption{QPS-add}
        \label{fig:sift1b2m-qps-add-1000}
    \end{subfigure}
    \hfill
    \begin{subfigure}{0.32\textwidth}
        \centering
        \includegraphics[width=\textwidth]{fig/experiment/sift1b_2m_QPS_delete_unit_1000.pdf}
        \caption{QPS-delete}
        \label{fig:sift1b2m-qps-delete-1000-supp}
    \end{subfigure}
    \hfill
    \begin{subfigure}{0.32\textwidth}
        \centering
        \includegraphics[width=\textwidth]{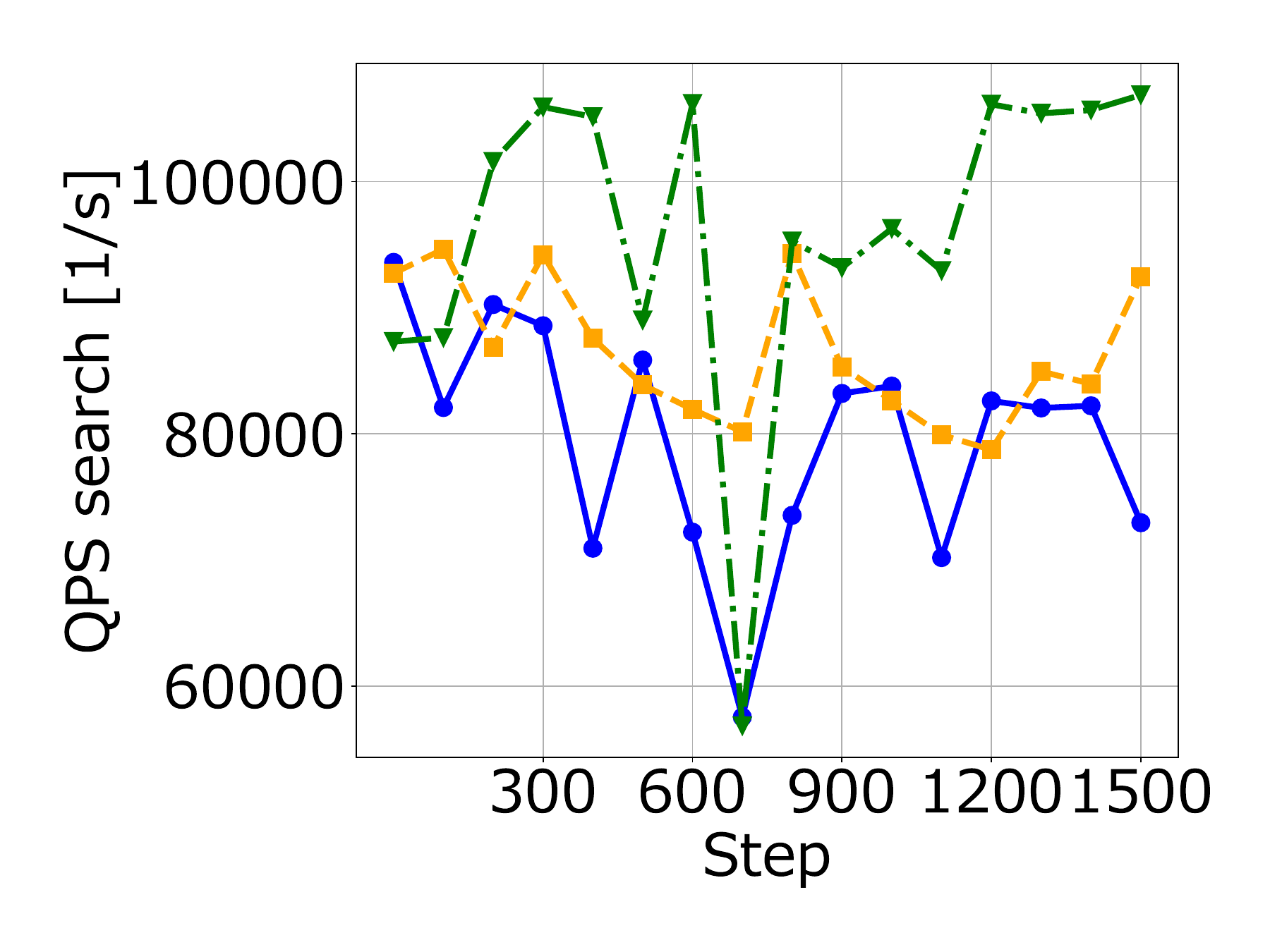}
        \caption{QPS-search}
        \label{fig:sift1b2m-qps-search-1000}
    \end{subfigure}
    \hfill
    \caption{Performance comparison of the three deletion methods at each step on SIFT1B with $b=10^3$.}
    \label{fig:sift1b2m-1000}
\end{figure*}

\begin{figure*}[tb]
    \centering
    \begin{subfigure}{0.32\textwidth}
        \centering
        \includegraphics[width=\textwidth]{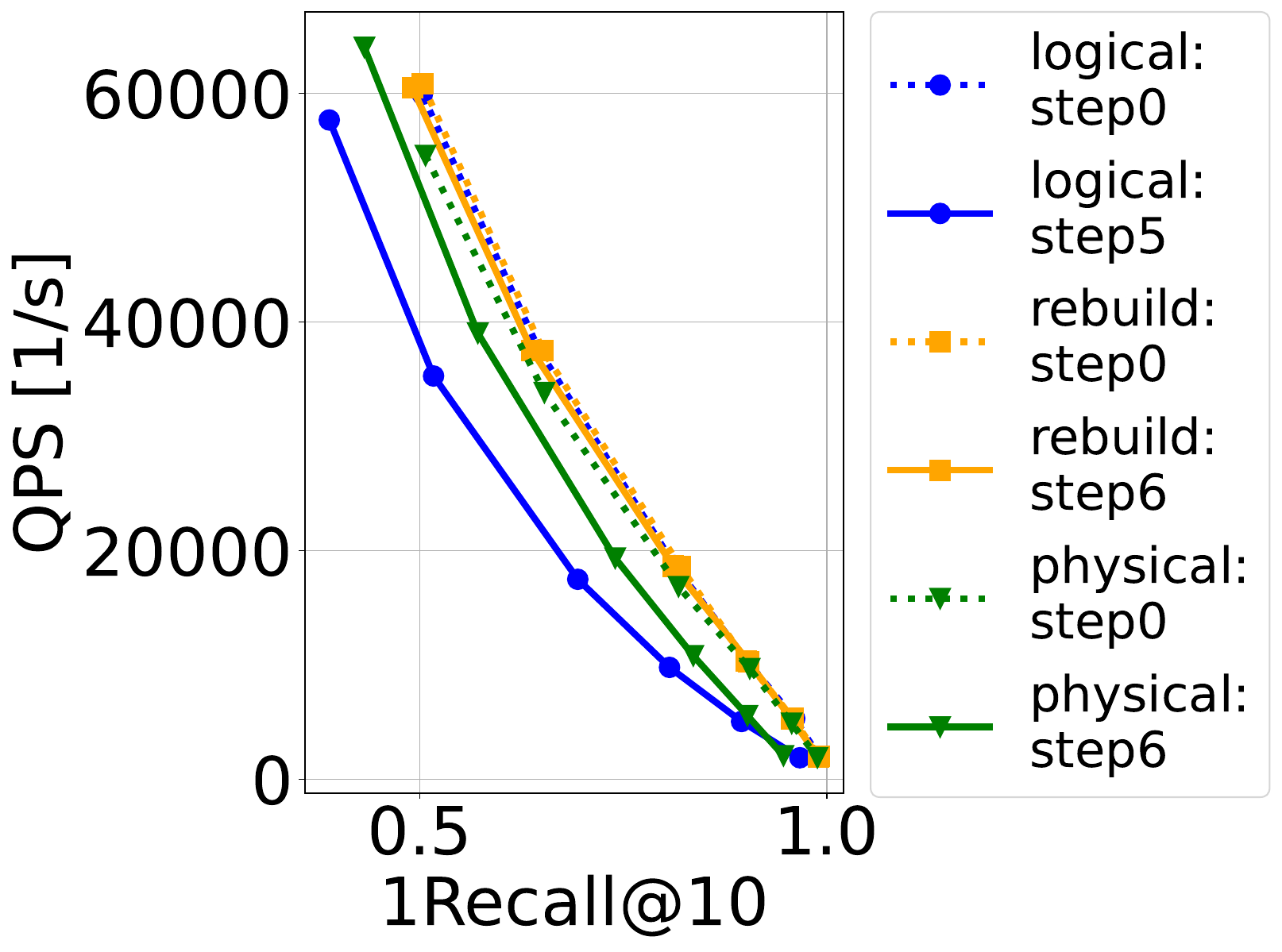}
        \caption{QPS-Recall}
        \label{fig:glove100-qps-recall}
    \end{subfigure}
    \hfill
    \begin{subfigure}{0.32\textwidth}
        \centering
        \includegraphics[width=\textwidth]{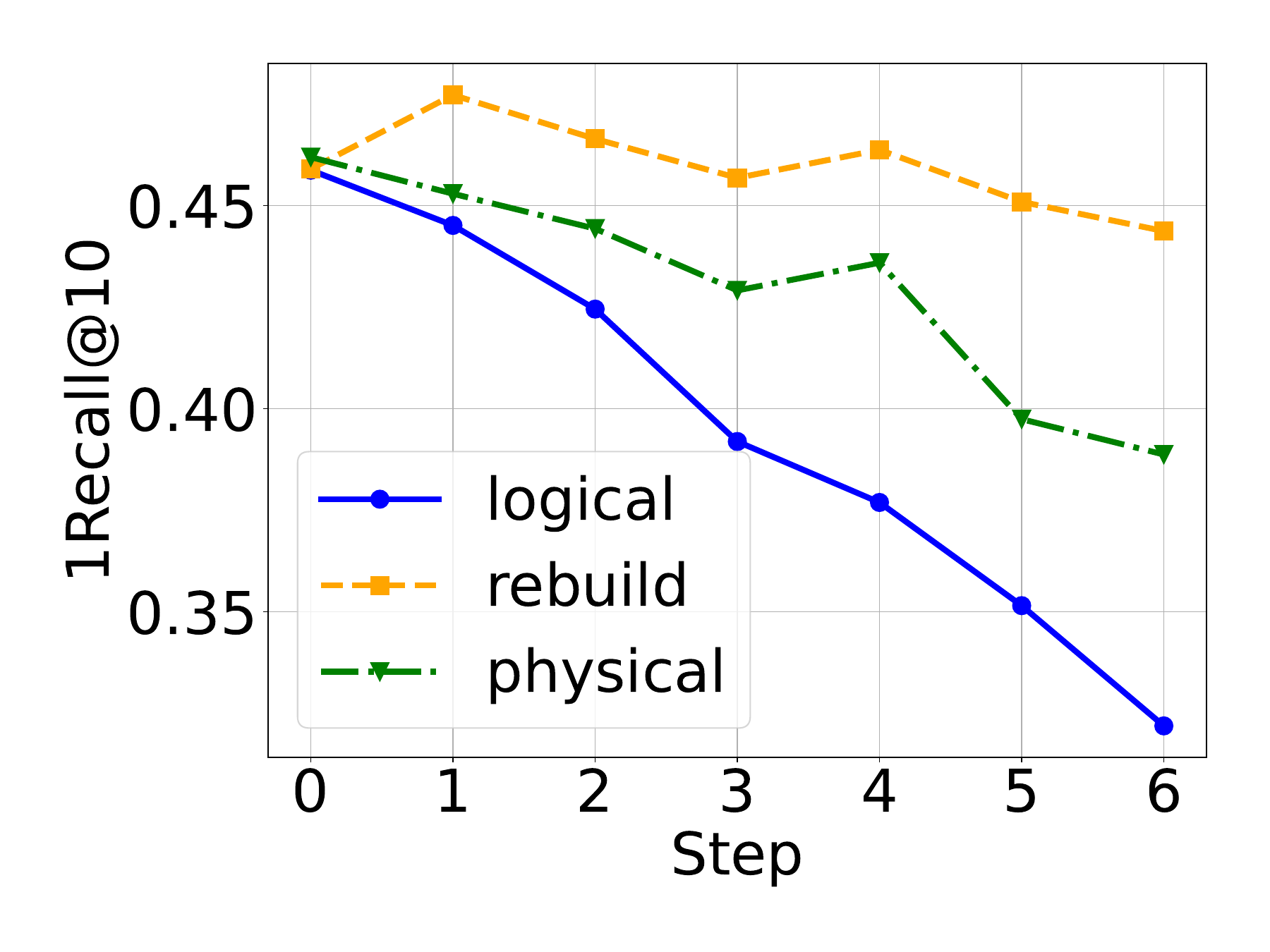}
        \caption{1-Recall@10}
        \label{fig:glove100-recall}
    \end{subfigure}
    \hfill
    \begin{subfigure}{0.32\textwidth}
        \centering
        \includegraphics[width=\textwidth]{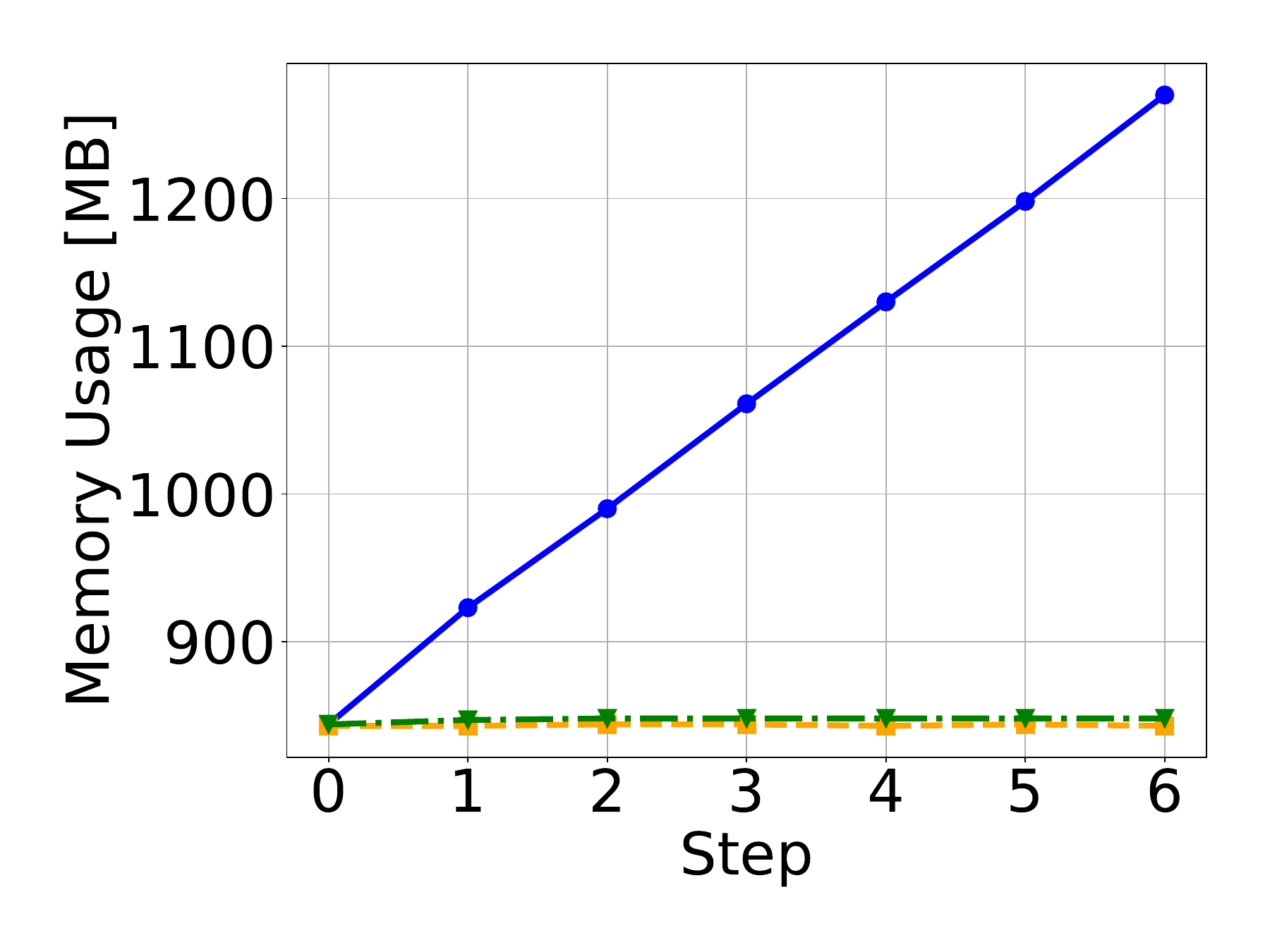}
        \caption{Memory Usage}
        \label{fig:glove100-memory}
    \end{subfigure}
    \hfill
    \begin{subfigure}{0.32\textwidth}
        \centering
        \includegraphics[width=\textwidth]{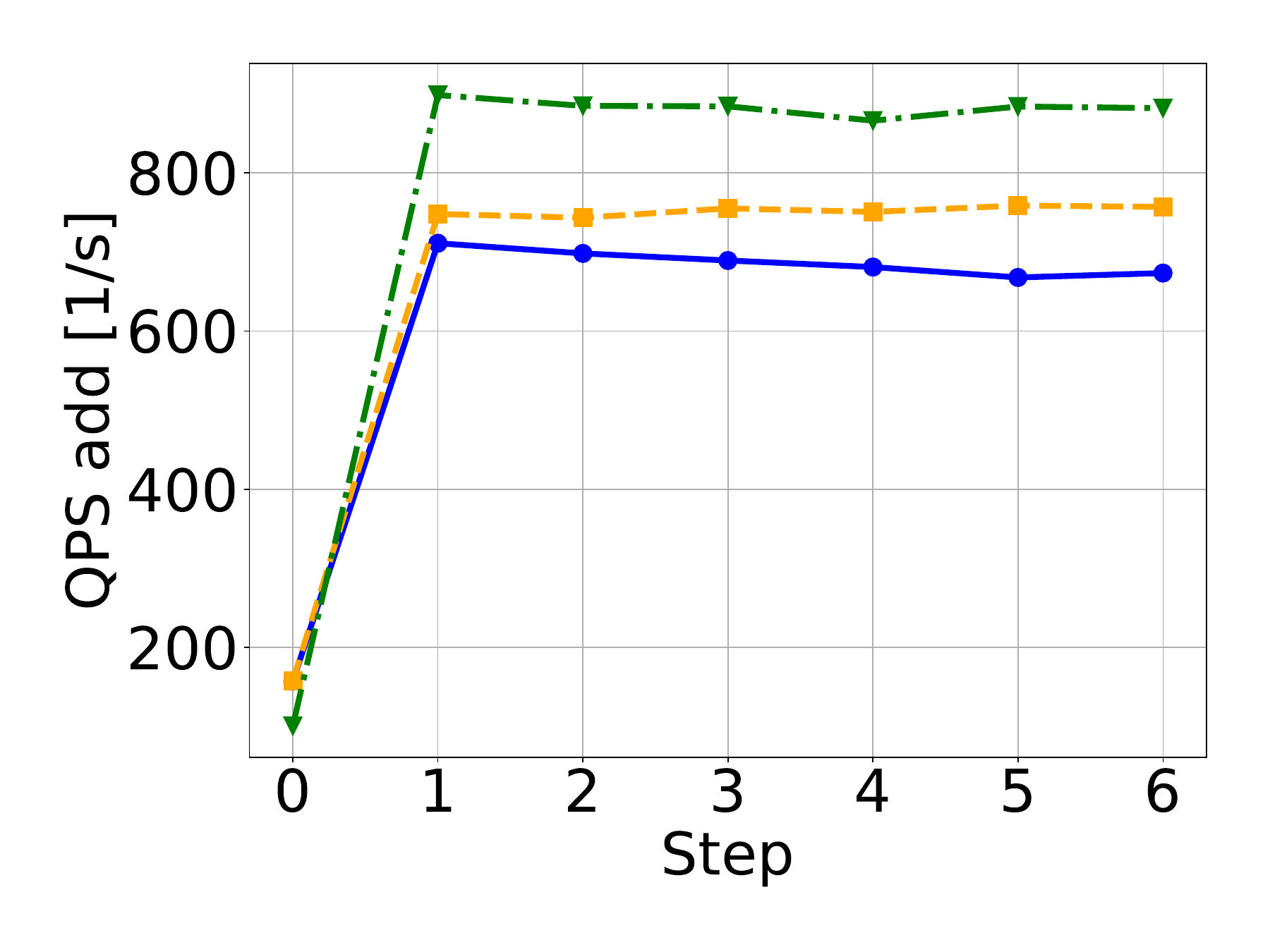}
        \caption{QPS-add}
        \label{fig:glove100-qps-add}
    \end{subfigure}
    \hfill
    \begin{subfigure}{0.32\textwidth}
        \centering
        \includegraphics[width=\textwidth]{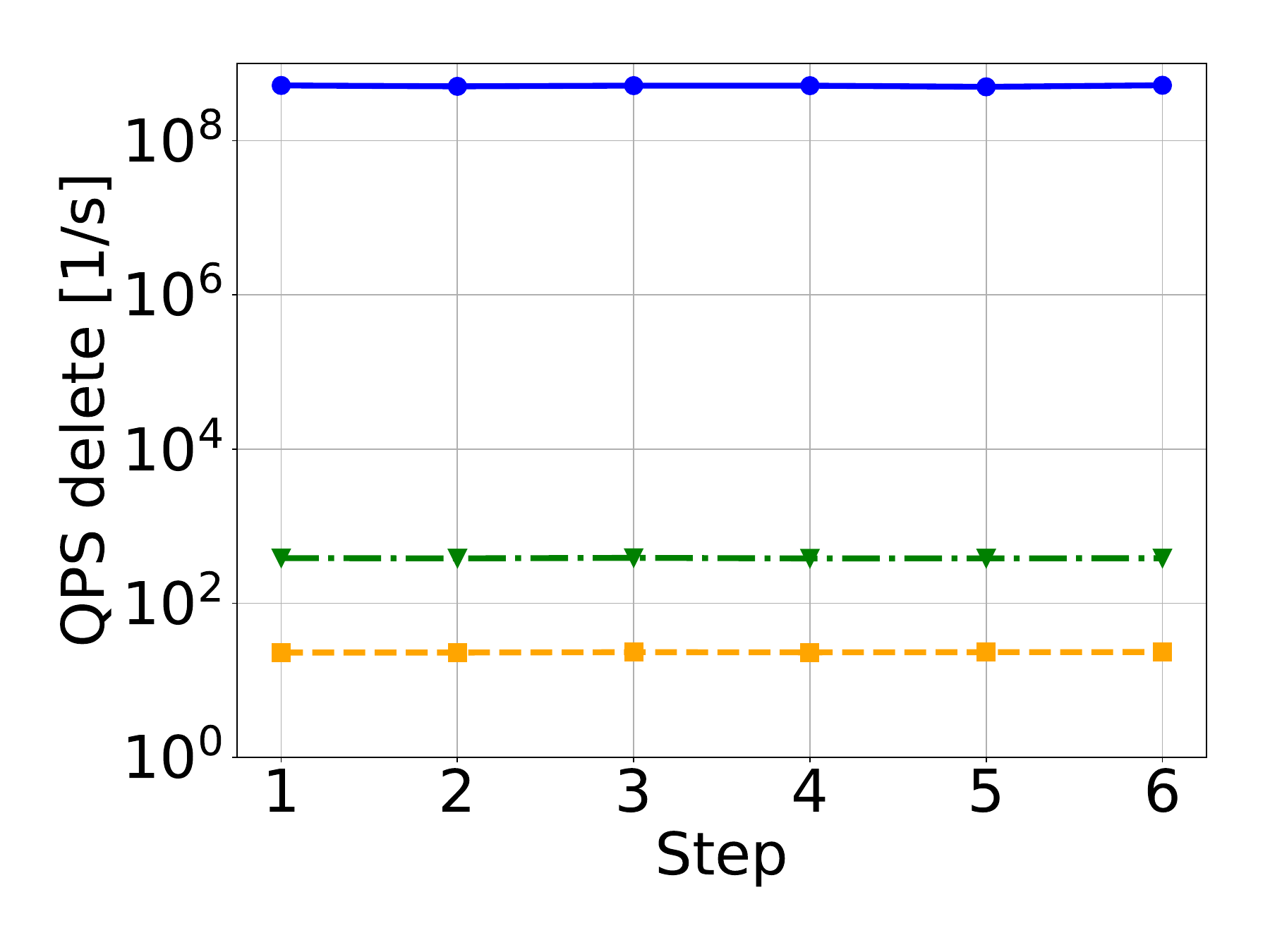}
        \caption{QPS-delete}
        \label{fig:glove100-qps-delete}
    \end{subfigure}
    \hfill
    \begin{subfigure}{0.32\textwidth}
        \centering
        \includegraphics[width=\textwidth]{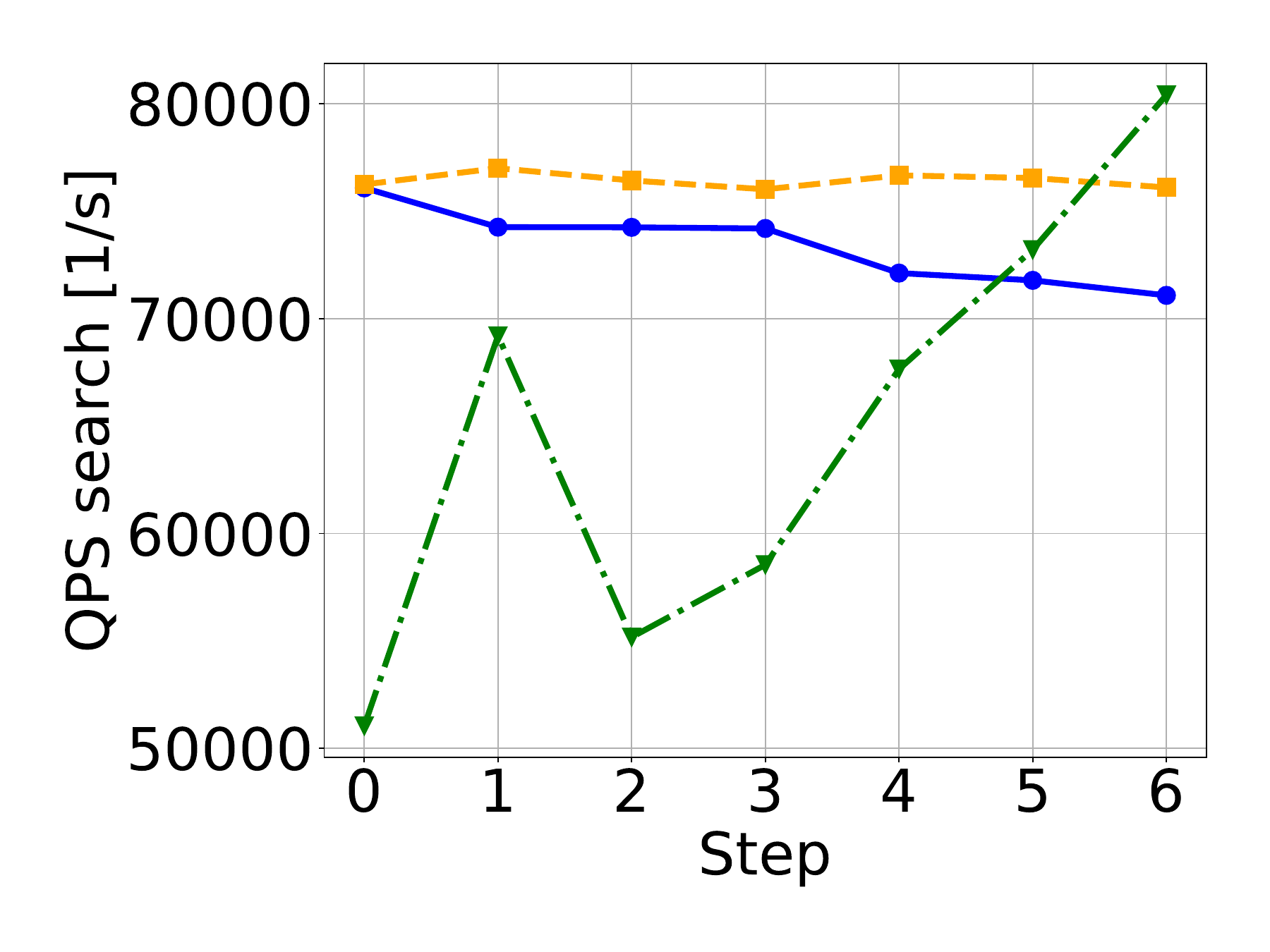}
        \caption{QPS-search}
        \label{fig:glove100-qps-search}
    \end{subfigure}
    \hfill
    \caption{Performance comparison of the three deletion methods at each step on GLOVE100.}
    \label{fig:glove100}
\end{figure*}

From \fig{\ref{fig:sift1m-qps-delete}}, \fig{\ref{fig:gist1m-qps-delete}}, \fig{\ref{fig:sift1b2m-qps-delete-100000-supp}}, \fig{\ref{fig:sift1b2m-qps-delete-1000-supp}}, \fig{\ref{fig:deep1m-qps-delete}} and \fig{\ref{fig:glove100-qps-delete}}, it is evident that across all datasets, logical deletion achieves the highest data deletion speed, on the order of approximately \( 10^{9} [1/{\rm{s}}] \). 
In contrast, both rebuilding and physical deletion operate at a significantly lower speed, at most on the order of \( 10^{3} [1/{\rm{s}}] \).  
When data insertion and deletion are performed in batches of \( b = 10^5 \), logical deletion can be completed in approximately \( 10^{-4} [{\rm{s}}] \), whereas physical deletion requires up to \( 10^{2} [{\rm{s}}] \).

From \fig{\ref{fig:sift1m-qps-delete}} and \fig{\ref{fig:gist1m-qps-delete}}, it can be observed that the dimensionality of the inserted and deleted vectors affects only the speed of rebuilding. SIFT1M~\cite{sift1m} has a dimensionality of 128, whereas GIST1M~\cite{sift1m} has a dimensionality of 960. This difference impacts rebuilding because it requires distance calculations during deletion. As the vector dimensionality increases, the time needed for a single-distance calculation also increases, leading to slower deletion speeds.

From \fig{\ref{fig:sift1b2m-qps-delete-100000-supp}} and \fig{\ref{fig:sift1b2m-qps-delete-1000-supp}}, it can be observed that the data deletion speed of physical deletion remains almost unchanged regardless of the batch size \( b \). This indicates that physical deletion primarily involves memory operations for the specified deletion queries, leading to a consistent processing speed.  
Specifically, when the batch size is reduced from \( b = 10^5 \) to \( b = 10^3 \), meaning the number of deletions per step is reduced to \( 1/100 \), the deletion speed of physical deletion remains nearly constant. In contrast, the speed of rebuilding decreases by approximately a factor of 100. This is because when the dataset size is relatively small, the processing time required for rebuilding remains almost constant, regardless of the number of deleted data points.

\subsection{QPS-search}\label{subsec:qps-search}
\begin{figure*}[tb]
    \centering
    \begin{subfigure}{0.32\textwidth}
        \centering
        \includegraphics[width=\textwidth]{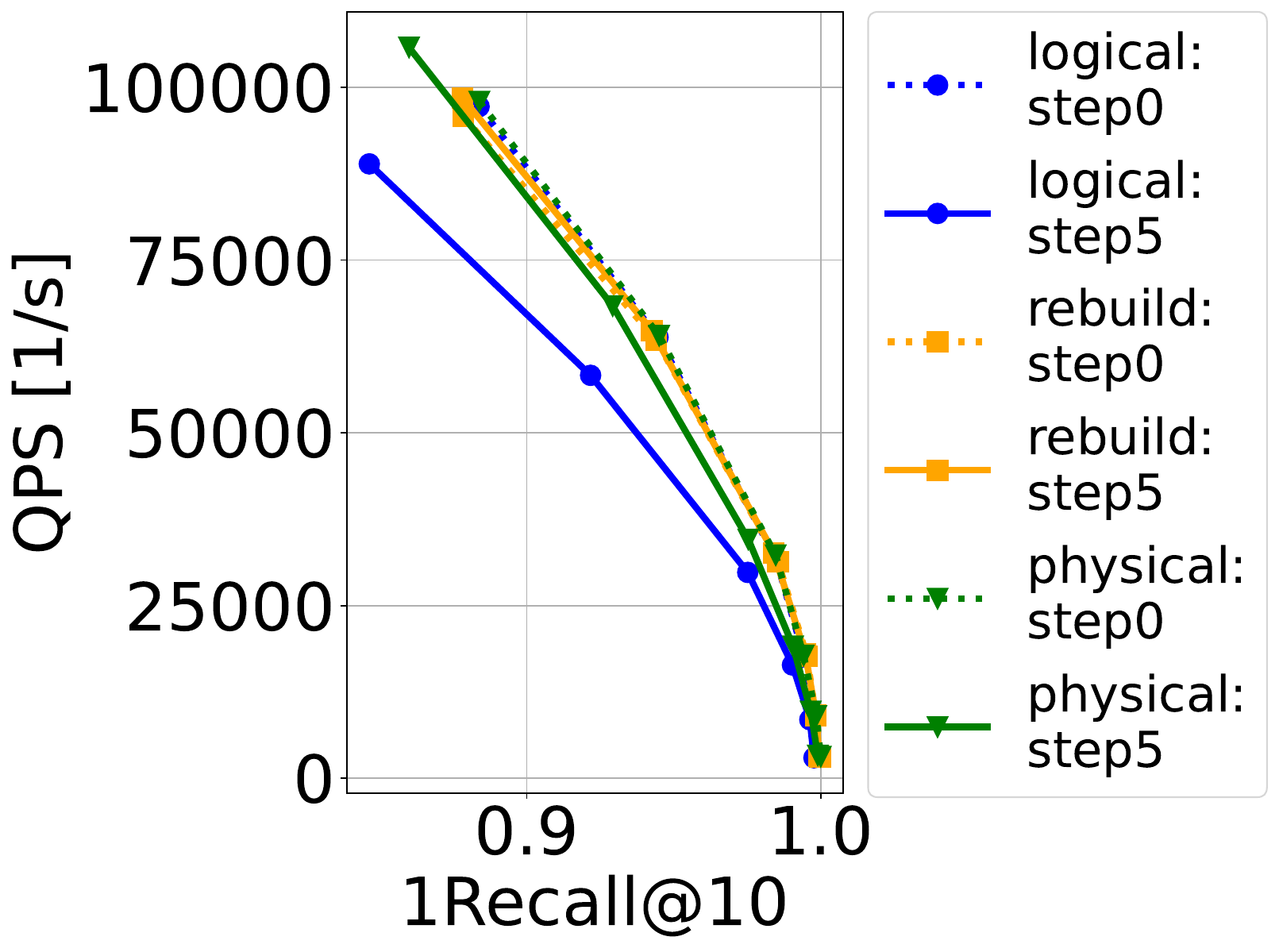}
        \caption{QPS-Recall}
        \label{fig:deep1m-qps-recall}
    \end{subfigure}
    \hfill
    \begin{subfigure}{0.32\textwidth}
        \centering
        \includegraphics[width=\textwidth]{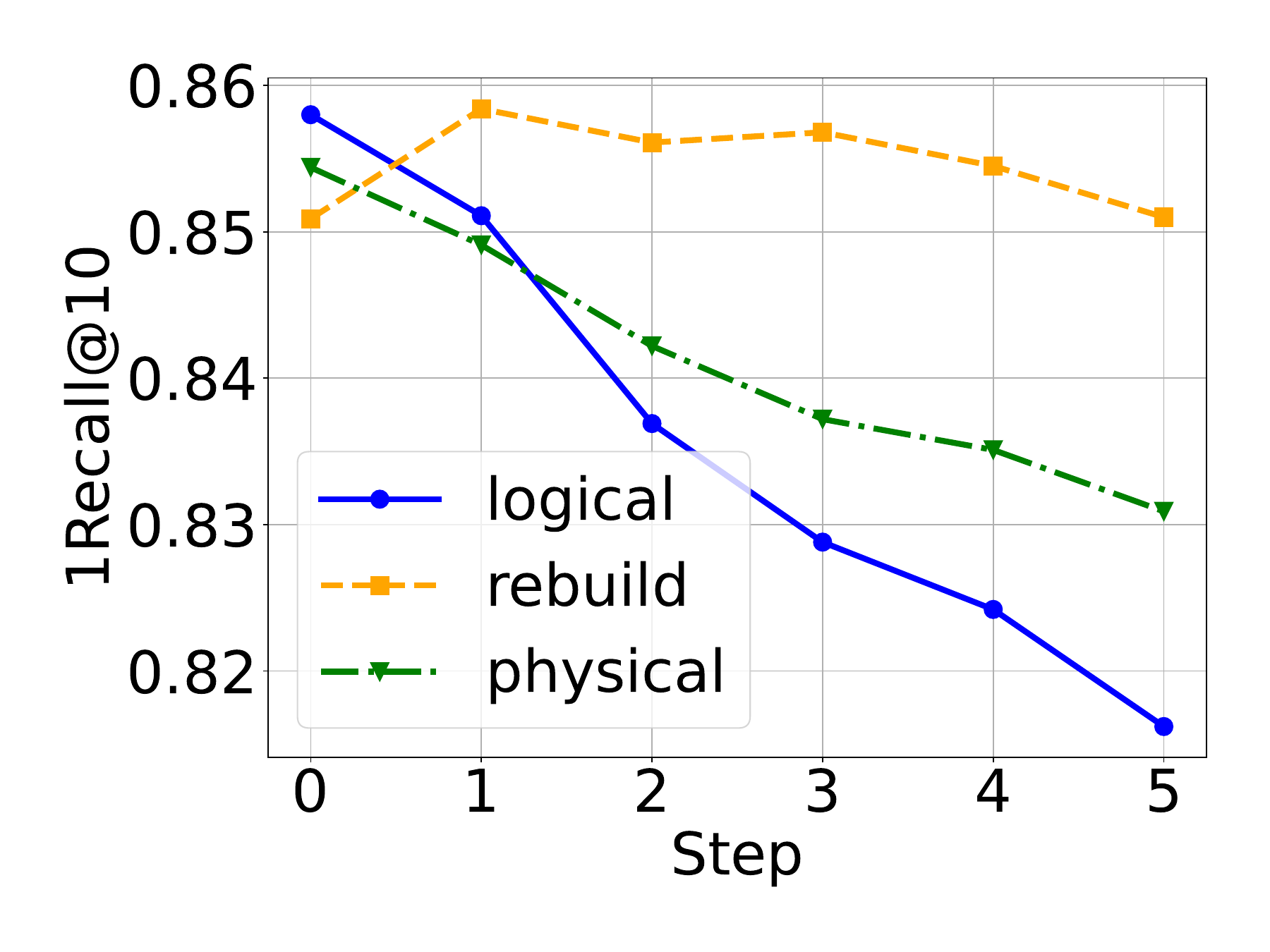}
        \caption{1-Recall@10}
        \label{fig:deep1m-recall}
    \end{subfigure}
    \hfill
    \begin{subfigure}{0.32\textwidth}
        \centering
        \includegraphics[width=\textwidth]{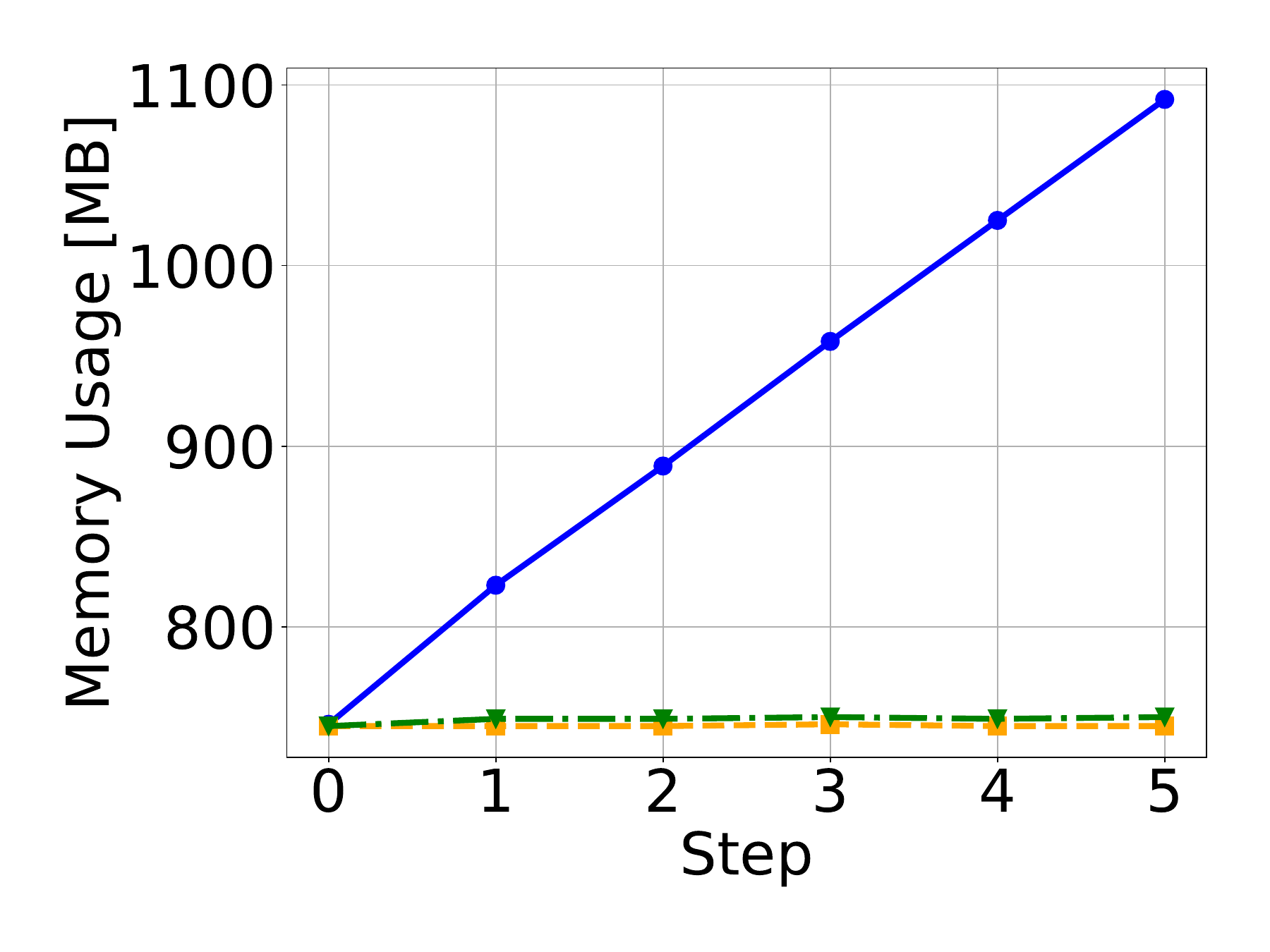}
        \caption{Memory Usage}
        \label{fig:deep1m-memory}
    \end{subfigure}
    \hfill
    \begin{subfigure}{0.32\textwidth}
        \centering
        \includegraphics[width=\textwidth]{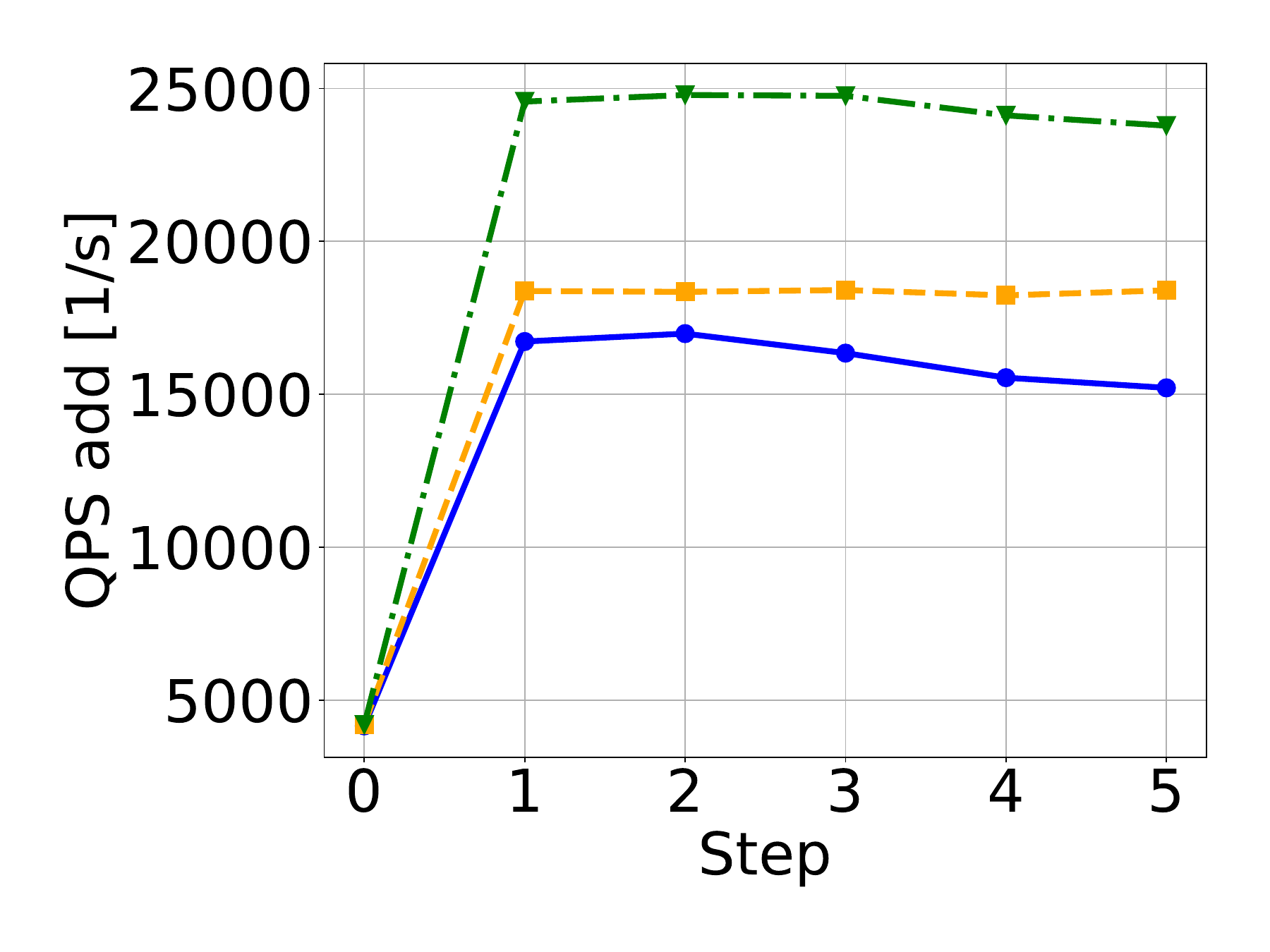}
        \caption{QPS-add}
        \label{fig:deep1m-qps-add}
    \end{subfigure}
    \hfill
    \begin{subfigure}{0.32\textwidth}
        \centering
        \includegraphics[width=\textwidth]{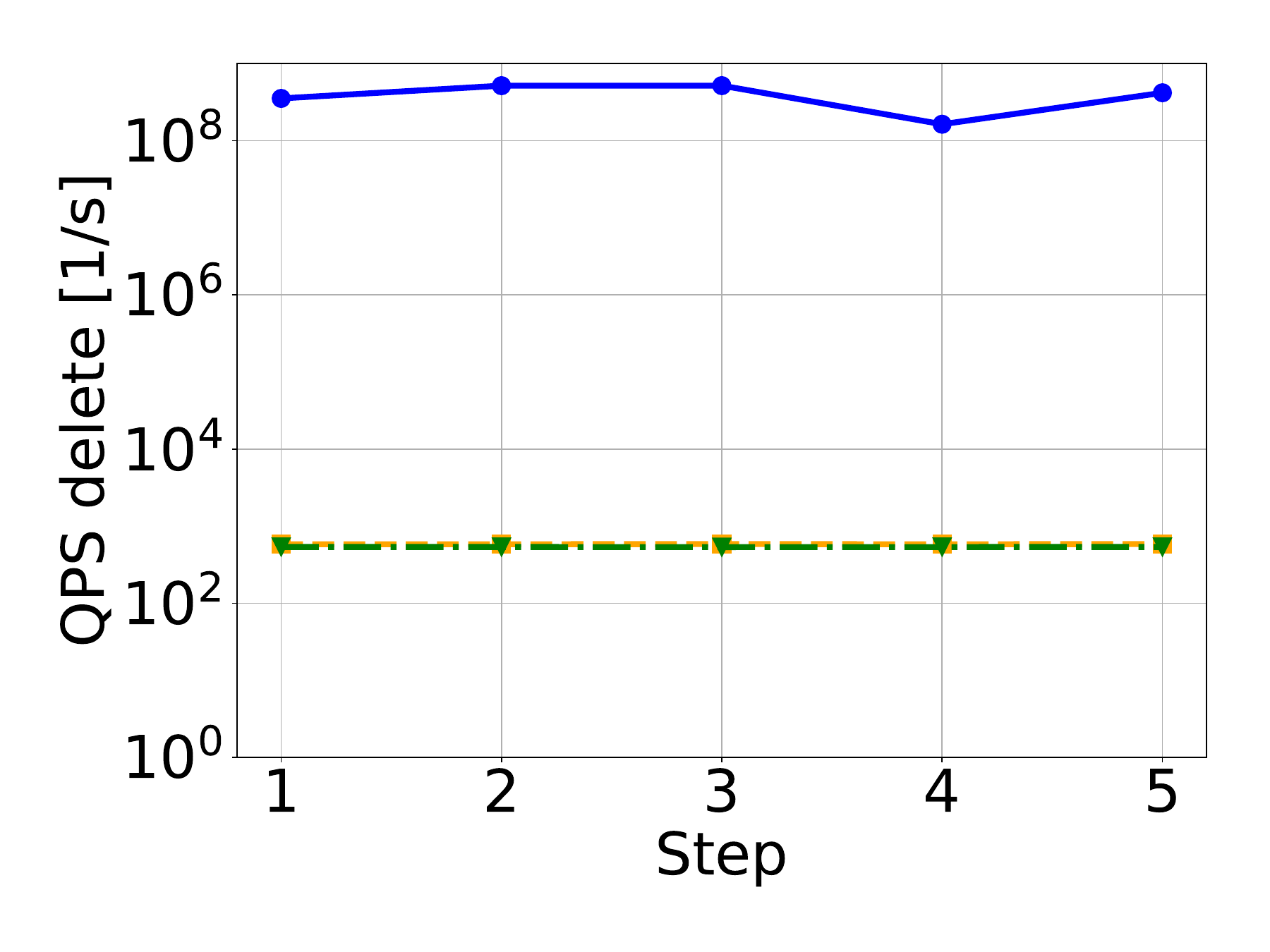}
        \caption{QPS-delete}
        \label{fig:deep1m-qps-delete}
    \end{subfigure}
    \hfill
    \begin{subfigure}{0.32\textwidth}
        \centering
        \includegraphics[width=\textwidth]{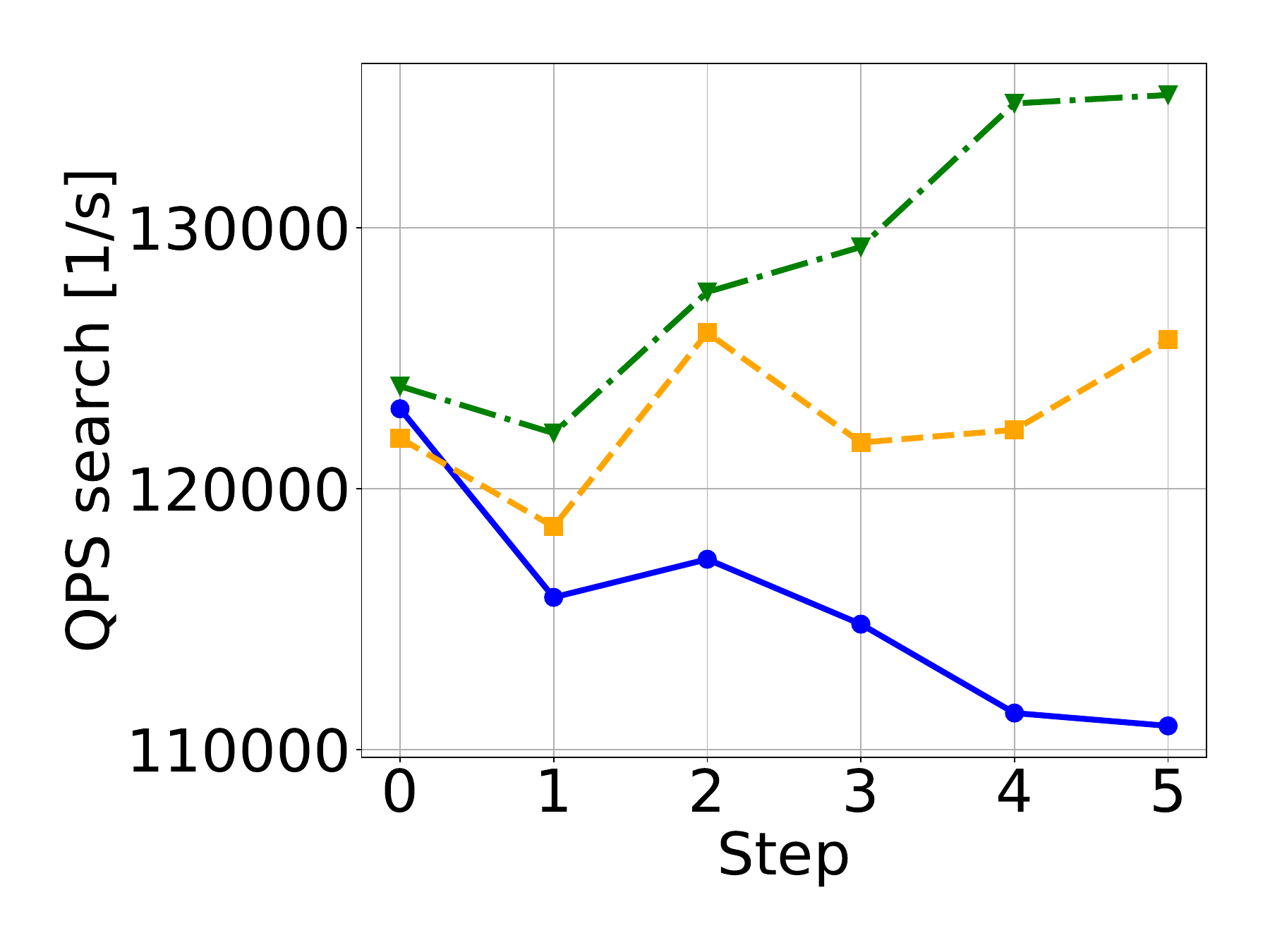}
        \caption{QPS-search}
        \label{fig:deep1m-qps-search}
    \end{subfigure}
    \hfill
    \caption{Performance comparison of the three deletion methods at each step on DEEP1M.}
    \label{fig:deep1m}
\end{figure*}

From \fig{\ref{fig:sift1m-qps-search}}, \fig{\ref{fig:gist1m-qps-search}}, \fig{\ref{fig:sift1b2m-qps-search-100000}}, \fig{\ref{fig:sift1b2m-qps-search-1000}} and \fig{\ref{fig:deep1m-qps-search}}, it is evident that across all datasets, search speed is highest when using physical deletion.  
As discussed in \scn{\ref{subsec:qps-add}}, this is likely because physical deletion gradually makes the graph sparser, reducing the number of distance calculations required during the search.

Similarly, across all datasets, logical deletion results in the slowest search speed. This is likely because, in logical deletion, an additional operation is required after the standard search process: the retrieved results must be filtered by referencing a flag array to exclude deleted data.

\section{Deletion Control} 
\label{sec:apx-deletion-control}

The algorithm of Deletion Control strategy is shown in \alg{\ref{alg:deletion-control}}.

\begin{algorithm}[tb]
\caption{Deletion Control}
\label{alg:deletion-control}
\DontPrintSemicolon
\SetKwInOut{Input}{Input}
\Input{required search accuracy $\alpha$; estimated $\theta$ and $\pi$.}
\uIf{$a \le \theta$}{
    \ForEach{update step}{
        \texttt{DELETE} via physical deletion;\;
    }
}
\Else{
    $s \leftarrow 0$\;
    \While{for each update step}{
        \texttt{DELETE} via logical deletion;\quad $s\leftarrow s+1$ \;
        \If{$s = \pi$}{
            \texttt{CONSTRUCT} to rebuild from current $\mathcal{P}$;\quad $s\leftarrow 0$ \;
        }
    }
}
\end{algorithm}

\clearpage

\clearpage

\clearpage

\clearpage
\newpage

\end{document}